\documentclass[10pt,twocolumn,letterpaper]{article}

\usepackage[pagenumbers]{cvpr} 

\definecolor{Coral}{rgb}{1, 0.47, 0.24}

\newcommand{\modelname}{Horizon-GS\xspace}

\usepackage{multirow}
\definecolor{cvprblue}{rgb}{0.21,0.49,0.74}
\usepackage{color}
\usepackage{xcolor}
\usepackage{colortbl}
\usepackage{marvosym}
\definecolor{tabfirst}{rgb}{1, 0.7, 0.7} 
\definecolor{tabsecond}{rgb}{1, 0.85, 0.7} 
\definecolor{tabthird}{rgb}{1, 1, 0.7} 
\usepackage[pagebackref,breaklinks,colorlinks,allcolors=cvprblue]{hyperref}

\def\paperID{2537} 
\def\confName{CVPR}
\def\confYear{2025}

\title{
    Horizon-GS: Unified 3D Gaussian Splatting for Large-Scale \\
    Aerial-to-Ground Scenes
}
\author{Lihan Jiang$^{1,3}$\thanks{*Equal contribution} \quad
Kerui Ren$^{2,3}$\protect\footnotemark[1] \quad
Mulin Yu$^{3}$ \quad
Linning Xu$^{4}$ \quad \\
Junting Dong$^{3}$ \quad
Tao Lu$^{5}$ \quad
Feng Zhao$^{1}$ \quad
Dahua Lin$^{4}$ \quad
\vspace{0.5em}Bo Dai$^{6,3}$\Letter \\
{\small $^1$University of Science and Technology of China, $^2$Shanghai Jiao Tong University,} \\
{\small$^3$Shanghai Artificial Intelligence Laboratory, $^4$The Chinese University of Hong Kong,} \\
{\small$^5$Brown University, $^6$The University of Hong Kong}
}

\begin{document}

\twocolumn[{
\renewcommand\twocolumn[1][]{#1}
\maketitle
\begin{center}
    \centering
    \vspace{-20pt}
    \captionsetup{type=figure}
    \includegraphics[width=1\textwidth]{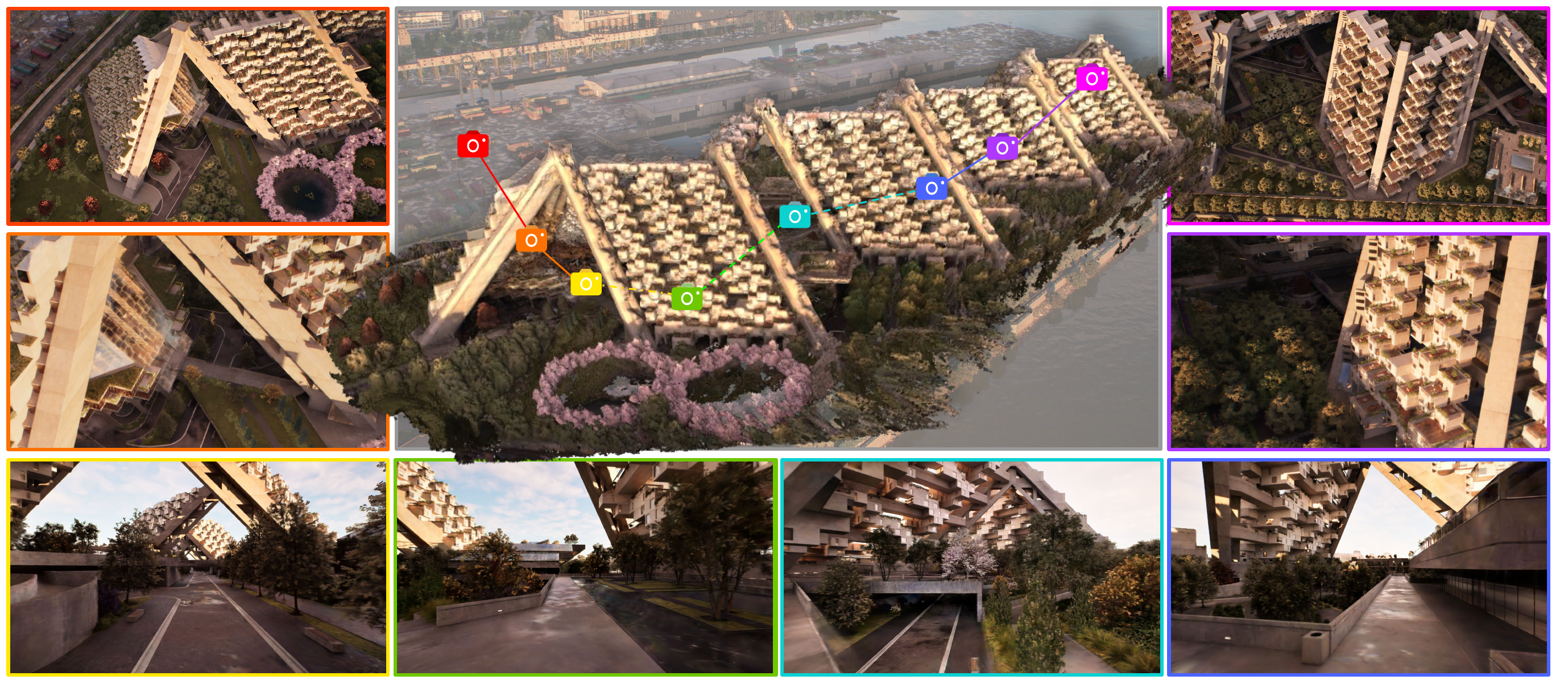}
    \vspace{-20pt}
    \captionof{figure}{\modelname enables high-quality rendering and reconstruction of aerial-to-ground scenes with unprecedented fidelity across scales, supporting drastic view changes. The colored camera trajectories in the center illustrate the novel viewpoints, while the reconstructed mesh is overlaid on the scene. The surrounding images show the corresponding predicted images for each viewpoint.
    }
    \label{fig:teaser}
    
\end{center}
}]

 \begin{abstract}
Seamless integration of both aerial and street view images remains a significant challenge in neural scene reconstruction and rendering. 
Existing methods predominantly focus on single domain, limiting their applications in immersive environments, which demand extensive free view exploration with large view changes both horizontally and vertically.
We introduce \modelname, a novel approach built upon Gaussian Splatting techniques, tackles the unified reconstruction and rendering for aerial and street views. 
Our method addresses the key challenges of combining these perspectives with a new training strategy, overcoming viewpoint discrepancies to generate high-fidelity scenes. 
We also curate a high-quality aerial-to-ground views dataset encompassing both synthetic and real-world scene to advance further research.
Experiments across diverse urban scene datasets confirm the effectiveness of our method.
Project page: \href{https://city-super.github.io/horizon-gs/}{\textcolor{magenta}{\textbf{https://city-super.github.io/horizon-gs/}}}.
\end{abstract}   
 \section{Introduction}
\label{sec:intro}






Modeling and rendering large-scale scenes has become essential across domains such as Embodied AI, digital twins, and autonomous driving. Neural radiance fields~\cite{mildenhall2021nerf} have revealed the potential to realize this vision. Recently, 3D Gaussian Splatting (3D-GS)~\cite{kerbl20233d} has advanced this field further, achieving exceptional visual quality alongside real-time rendering speeds. These capabilities markedly accelerate the transformation of the physical world into digital spaces, supporting applications from real-to-sim simulation and autonomous navigation to immersive VR/AR experiences and the metaverse.
%

Despite recent advancements in rendering aerial~\cite{lin2024vastgaussian,liu2024citygaussian,chen2024dogaussian} and street views~\cite{kerbl2024hierarchical} with 3D-GS, existing methods remain limited to training on a single view type, resulting in a disjointed experience. We aim to bridge this gap by integrating both perspectives, enabling a seamless and unified experience across downstream applications.
%
%
%
Our preliminary experiments on state-of-the-art Gaussian Splatting methods reveal two primary challenges in integrating aerial and street views: 1) conflicts in Gaussian densification arising from the substantial disparity between aerial and street views, and 2) an uneven distribution of camera data, with a bias toward street views. During the densification process, limited visibility from street views impedes consistent gradient accumulation, thereby disrupting the Gaussian instantiation process across the scene. Moreover, the variation in detail between aerial and street views makes a single static set of Gaussian primitives insufficiently flexible, restricting smooth transitions between viewpoints. 


%

To address above challenges and achieve a seamless integration for both aerial and street views, we propose a novel method, \modelname,  that overcomes the limitations of existing approaches. 
Specifically, we propose a coarse-to-fine training strategy that explicitly divides the entire training process into two stages: the first stage benefits more from aerial images to establish a coarse geometry, and the second stage focuses on adding finer details mainly from street views. Combined with a camera distribution balance strategy, our method achieves state-of-the-art quality for both aerial and street views, delivering superior rendering quality and geometry accuracy. Moreover, our approach is easily adaptable for large-scale reconstruction. 
We also notice that, the lack of calibrated aerial and street datasets is a critical issue that hinders unified model training. To advance the field and evaluate our method, we construct a large-scale dataset consisting of five synthetic and two real-world scenes, encompassing both aerial and street views.

In summary, the main contributions of our method are:
\begin{itemize} 
\item We tackle the challenging and important task of unified large-scale scene reconstruction from both aerial and street views, starting with an in-depth analysis of the core conflicts that hinder integration across these perspectives.
\item We propose an efficient framework that addresses the inherent discrepancies and unifies the contribution from aerial and street views. Our framework is also adaptable to both 3D Gaussians for realistic novel view synthesis and 2D Gaussians for accurate geometry reconstruction.
\item We present a high-quality and diverse dataset that includes both synthetic and real-world data, specifically curated to support balanced, cross-view reconstruction.
\item Our approach achieves state-of-the-art performance in small-scale and large-scale scene reconstruction, setting a new benchmark for unified aerial and street modeling.
\end{itemize}

 \section{Related work}
\label{sec:related_work}


\paragraph{Neural Scene Representations.}
Neural Radiance Fields (NeRF)~\cite{mildenhall2021nerf} have sparked significantly interest in novel view synthesis due to their photorealistic visual quality. 
Recent research~\cite{liu2020neural, fridovich2022plenoxels,sun2022direct,chen2022tensorf,muller2022instant,xiangli2023assetfield} has introduced various hybrid grid representations, reducing the number of sampling points and MLP layers, accelerating both the training and rendering processes.
More recently, 3D Gaussian Splatting (3D-GS)~\cite{kerbl20233d} has achieved more detailed visual quality with efficient training times by using anisotropic 3D Gaussians, while enabling real-time rendering through tile-based splatting methods. 
It quickly revolutionizes neural rendering and is widely applied in various fields, especially in many large-scale applications, such as autonomous driving~\cite{chen2023periodic,zhou2024drivinggaussian,yan2024street,wei2024editable}, geometry reconstruction~\cite{yu2024gsdf,yu2024gaussian,guedon2024sugar}, and 3D generative tasks~\cite{tang2023dreamgaussian,tang2024lgm,xu2024grm,zhang2025gs}.

\paragraph{Large-scale Scene Modeling.}
NeRF and 3D Gaussian Splatting have sparked a new trend in neural modeling for large-scale scenes, pushing the boundaries of visual fidelity and real-time performance. Extensive neural representation methods have been developed for large-scale city reconstruction and rendering.
However, most methods target either aerial~\cite{xiangli2022bungeenerf,turki2022mega,xu2023grid,zhenxing2022switch,lin2024vastgaussian,liu2024citygaussian,chen2024dogaussian,ren2024octree} or street views~\cite{rematas2022urban,tancik2022block, liu2023real, kerbl2024hierarchical}. For aerial views, models like VastGaussian~\cite{lin2024vastgaussian} and CityGaussian~\cite{liu2024citygaussian} use data partitioning for parallel training, while DoGaussian~\cite{chen2024dogaussian} employs a global node to ensure consistent inference. For street scenes, Hierarchical-3DGS~\cite{kerbl2024hierarchical} optimizes and renders Gaussians in an LOD hierarchy. However, these methods do not naturally support a unified reconstruction from both aerial and street views.
The work closest to ours is UC-GS~\cite{zhang2024drone}, which incorporates cross-view uncertainty to enhance details in extrapolated views but lacks the scalability for large-scale applications, limiting its practicality.

Another major barrier to unified aerial and street modeling is the scarcity of datasets with calibrated images from both perspectives. Current large-scale neural rendering datasets, such as UrbanScene3D~\cite{lin2022capturing}, Mill19~\cite{turki2022mega}, Quad 6K~\cite{crandall2012sfm}, and MatrixCity~\cite{li2023matrixcity}, Waymo Block-NeRF~\cite{tancik2022block}, KITTI-360~\cite{liao2022kitti}, and Hierarchical-3DGS~\cite{kerbl2024hierarchical}, typically cater to a single view type. MatrixCity is the only dataset with both views, but it is synthetic and lacks diversity. To fill this gap, we introduce a new dataset with a variety of synthetic and real-world scenes, supporting research on large-scale, unified reconstruction.

\section{\modelname}
\label{sec:method}

\begin{figure*}[t!]
    \centering
    \includegraphics[width=1\linewidth]{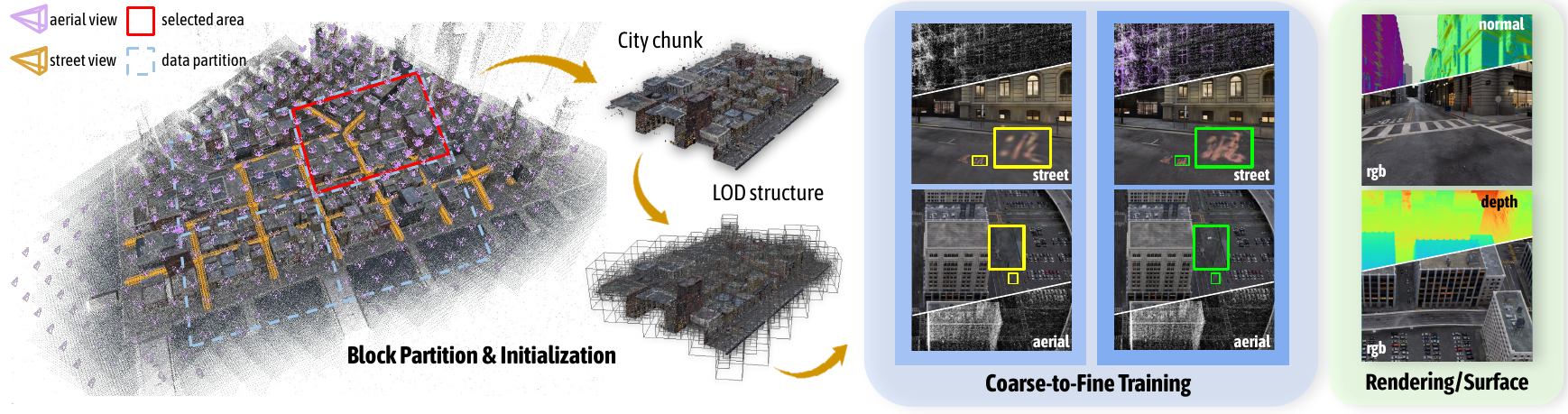}
    \caption{\textbf{Pipeline of \modelname.} We divide large-scale scenes into chunks. For each chunk, we initialize LOD-structured anchors and conduct the coarse-to-fine training process. Specifically, the coarse stage reconstructs the overall scene, while the fine stage enhances street view details (highlighted in \textcolor[RGB]{192, 113, 255}{purple}). 
    We can derive RGB, depth, and normal images by utilizing different primitive attributes (2D/3D Gaussians) with a single shared underlying structure.}
    \vspace{-7pt}
    \label{fig: pipeline}
\end{figure*}

\begin{figure}[htbp]
\centering
    \includegraphics[width=\linewidth]{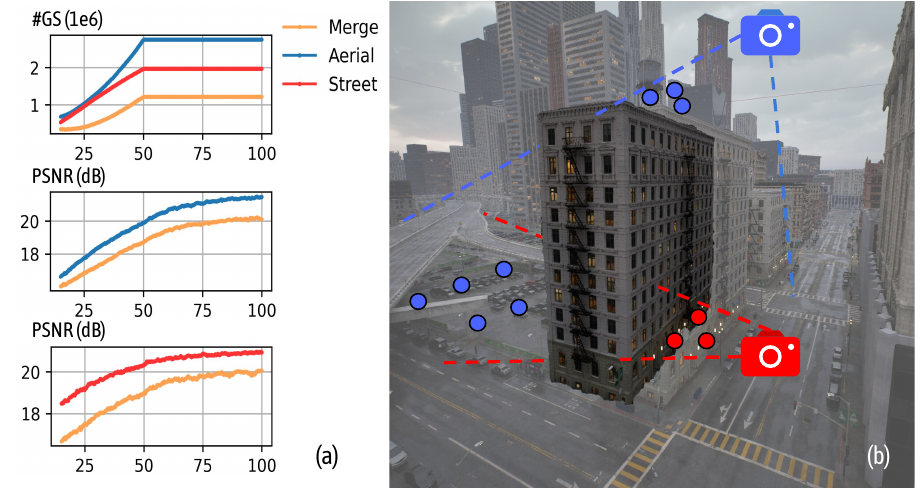}
    \caption{
    (a) Test curves for PSNR and the number of Gaussian primitives across aerial only, street only, and merged views from 15k to 100k iterations on our proposed Road scene. (b) Gradient conflicts restrict the optimization of Gaussian primitives because street views tend to exclude \textcolor[RGB]{0, 0, 255}{blue} Gaussian primitives due to their lower contribution, while aerial views do the opposite.
    }
    \vspace{-7pt}
    \label{fig:challenge}
\end{figure}


While substantial progress has been made in reconstructing urban scenes from either aerial or street images~\cite{lin2024vastgaussian,liu2024citygaussian,chen2024dogaussian,kerbl2024hierarchical}, the challenge of combining both perspectives into a unified model remains unsolved and largely overlooked. This task is non-trivial, requiring solutions to bridge the significant disparities in scale, perspective, and appearance between aerial and street views. In this section, we first analyze the core challenges in Section~\ref{sec: challenges}, briefly describe the base modules in Section~\ref{sec:3dgs_variants}, and then detail the proposed framework that leverages a two-stage training strategy and the enhanced 3D Gaussian densification operators in Section~\ref{sec: training_strategy}. Additionally, how to apply the framework to large-scale scenes is discussed in Section~\ref{sec:urban-strategy}. Finally, we describe the loss functions and regularization in Section~\ref{sec:loss}.

\subsection{Challenges}
\label{sec: challenges}
Before presenting our framework, we first analyze the inherent conflicts in using 3D Gaussians for simultaneous training on aerial and street views. We apply the state-of-the-art Scaffold-GS~\cite{lu2023scaffold} on our collected real-world scene which features both aerial and street views. Naturally, we aim to combine these two classes of views during reconstruction, as they complement each other. But we notice that reconstructing either aerial or street views alone has achieved much better results than training them together, as shown in Fig.~\ref{fig: pipeline} (b) and Fig.~\ref{fig:challenge}. Combining these perspectives presents two major challenges: (1) conflicting gradient accumulation and (2) disparity in view information coverage.

The first challenge arises from the differences between aerial and street views, which lead to conflicting gradient updates during training. While street views cause the densification policy to remove irrelevant, occluded Gaussians, aerial views promote regrowth in these regions. This interference disrupts the Gaussian densification process, making it unreliable and ultimately degrading reconstruction quality. Fig.~\ref{fig:challenge} confirms that joint training across both domains consistently underperformed separate training, even with extensive iterations, and with more observed views, the number of final Gaussian primitives are even less.

The second challenge is the imbalance in camera coverage: street views capture detailed, nearly 360-degree scenes, while aerial views are downward-focused and less frequent. This uneven distribution shifts the reconstruction focus toward street details, preventing a balanced integration of both perspectives. Together, these challenges emphasize the complexity of unified aerial-street modeling and the necessity for specialized solutions.

\subsection{Gaussian Splatting Base Modules} 
\label{sec:3dgs_variants}
3D Gaussian Splatting (3D-GS)~\cite{kerbl20233d} represents scenes using anisotropic 3D Gaussian primitives, achieving high rendering quality at significantly faster speeds. Each Gaussian $G$ is defined by parameters: a center $\mu \in \mathbb{R}^3$, a rotation quaternion $q \in \mathbb{R}^4$, scaling vector $s \in \mathbb{R}^3$, opacity $\sigma \in [0,1]$, and a color feature $c$ either encoded with spherical harmonics or direct RGB feature for view-dependent/independent color. To render images, the 3D Gaussians are projected onto the image plane as 2D Gaussians $G^{\prime}(x)$, ordered front-to-back, and $\alpha$-blended for pixel color calculation.

Scaffold-GS~\cite{lu2023scaffold} further introduces anchors to reduce redundancy and enhance storage efficiency. From each anchor, $k$ neural Gaussians are emitted, with centers determined by learnable offsets. Properties are decoded by MLPs from anchor features and viewing positions, creating a more robust and compact representation.

2D Gaussian Splatting (2D-GS)~\cite{huang20242d} collapses 3D Gaussians into 2D oriented planar Gaussian disks for surface accuracy. An adaptive rasterizer efficiently renders these 2D Gaussians, and depth distortion and normal consistency losses are added to improve reconstruction quality. Finally, a mesh is extracted using a truncated signed distance function (TSDF), which fuses multi-view depth maps from the optimized 2D-GS field.

For enhanced robustness, we adopt the anchor design from Scaffold-GS~\cite{lu2023scaffold}, incorporating a flexible neural Gaussian representation. 
Specifically, the properties of different neural Gaussians, both 3D and 2D, can be generated by learnable MLPs. We utilize neural 3D Gaussians to improve rendering quality and neural 2D Gaussians for high-fidelity geometry reconstruction.

\subsection{Aerial-Street Joint Reconstruction}
\label{sec: training_strategy}

\paragraph{Coarse-to-Fine Training.}

The significant difference in aerial and street views lead to performance challenges when training them together, necessitating separate handling for each modality. 
Inspired by BungeeNeRF~\cite{xiangli2022bungeenerf}, we propose a two-stage coarse-to-fine training strategy where street views refine the details of aerial views. In the first stage, we focus primarily on the aerial views to establish a coarse geometry, accumulating gradients for densification from the aerial images alone and allowing the Gaussian primitives to fully develop and cover a broader feature space. Street views guide the initial placement of fine-grained 3D Gaussians, ensuring their alignment with the global scene. In the second stage, we freeze the MLP weights and attributes of the Gaussian primitives from the first stage to preserve the skeleton structure. Street views then play a key role in refining this skeleton, adding finer details and achieving a balanced integration of both views while efficiently resolving their inherent conflicts.

We modify the densification policy in the second stage to enhance the capture of fine-grained details in large-scale scenes with sparse images. The original Scaffold-GS method averages screen-space gradients of neural Gaussians within each voxel. While effective for object-centric scenes, it struggles to model local details in sparsely observed areas. As a result, subtle features remain under-optimized, leading to blurred renderings.

Here we consider neural Gaussians with higher gradients, greater opacity, and larger projected screen size as more significant, inspired by Hierarchical-3DGS~\cite{kerbl2024hierarchical}.
Specifically, we compute the max gradients $\nabla_g$, average opacity $\sigma$ and max projected radius $r$ of the included neural Gaussians for each $N$ iterations. Those neural Gaussians satisfying:
\begin{equation}
\label{eq:den_policy}
    \nabla_g \cdot r \cdot \sigma^{\tau_{\sigma}} > \tau_g,
\end{equation}
where $\tau_{\sigma}$ and $\tau_g$ are pre-defined thresholds, are treated as significant and used as new anchors.

\paragraph{Balanced Camera Distribution.}
\label{sec: camera_balance}
As previously mentioned, the imbalanced distribution of cameras can lead to 3D-GS overfitting to street view details during joint training, causing blurring and loss of structural detail in aerial views.
Inspired by NeRF Director~\cite{xiao2024nerf}, we emphasize the importance of view selection and introduce a straightforward yet effective mechanism for balancing camera distribution. Specifically, we randomly select a value from the interval $[0,1]$. If the value is smaller than $\frac{R}{R+1}$, an aerial image is chosen for training; otherwise, a street image is picked. In detail, we set $R=2$ in the first stage to ensure thorough training of the aerial views and reduce it to 1 in the second stage.

\paragraph{Multi-resolution LOD Construction.}
Considering the differing levels of detail in aerial and street view images, and to enable real-time rendering, we adopt a Level-of-Detail (LOD) strategy, inspired by Octree-GS~\cite{ren2024octree}. Specifically, for content with varying details between aerial and street views, we set different LOD levels: $K_\text{aerial}$ for aerial views and $K$ for the whole views, corresponding to the two-phase training process.
\begin{equation}
    \begin{aligned}
        K_\text{aerial} &= \lfloor \log_{2}(D_\text{aerial}/d_\text{aerial}) \rfloor + 1, \\
        K &= K_\text{aerial} + \lfloor \log_{2}({d}_\text{aerial}/{d}_\text{street}) \rfloor.
    \end{aligned}
\end{equation}
where $D_\text{aerial}$ and $d_\text{aerial}$ are the $r_d$-th largest and $r_d$-th smallest aerial distances, while ${d}_\text{street}$ is the $r_d$-th smallest street distance. Note that, our model is set to $K$ layers and $K_\text{aerial}$ is only used in the first training stage. Specifically, in the first stage, 
we only activate $K_\text{aerial}$ layers and add new anchors to the same LOD level as the significant source anchors. In the second stage, we open all layers and add the new anchors to the next LOD level to capture the higher-frequency information of street views.




\begin{figure*}[htbp]
    \centering
    \includegraphics[width=\linewidth]{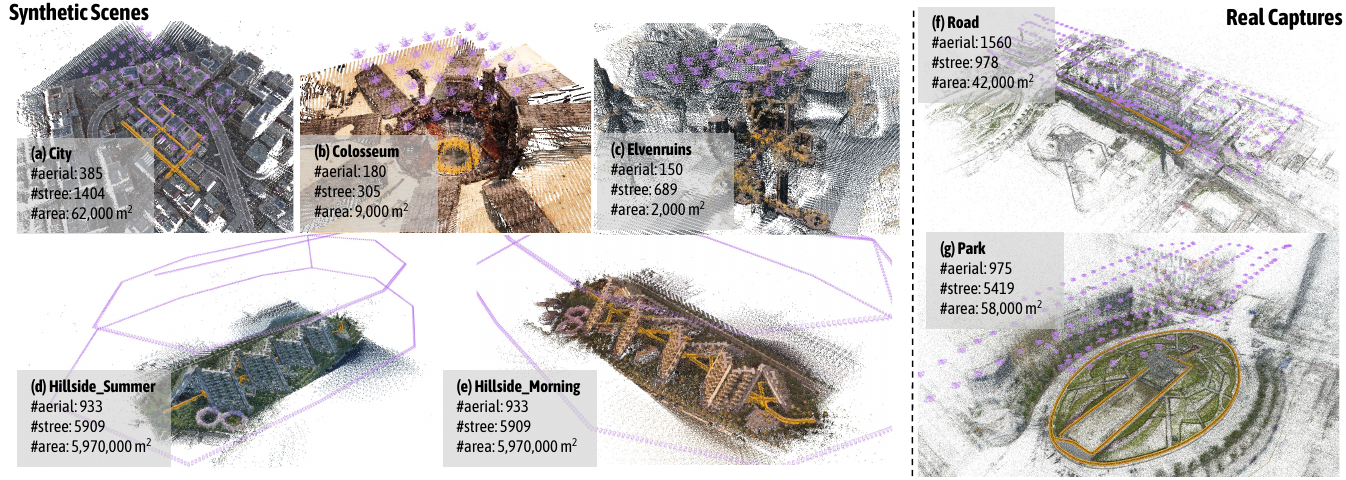}
    \vspace{-0.7cm}
    \caption{\textbf{Visualization of our constructed dataset.} All the $7$ scenes contain calibrated aerial and street view images. We illustrate the scenes with the point clouds and the corresponding image capture poses. The trajectory of aerial views is shown in \textcolor[RGB]{192, 113, 255}{purple}, while street views are represented in \textcolor[RGB]{250, 176, 26}{yellow}. Our dataset contains 5 synthetic scenes (a-e) and 2 real scenes (f-g).}
    \vspace{-5pt}
    \label{fig:data_vis}
\end{figure*}

\subsection{Large-scale Scene Training}
\label{sec:urban-strategy}
In this section, we present a partitioning strategy to adapt our framework for large-scale urban scenes, supporting both model scalability and rendering fidelity.


Gaussian-based methods often suffer from blurring artifacts when reconstructing large-scale scenes, mainly due to challenges in densifying extensive image sets that require fine-tuned parameters. Additionally, limited GPU memory presents a hurdle when training on large-scale scenes. A common solution is the divide-and-conquer strategy, where a whole scene is split into smaller chunks. For instance, VastGaussian \cite{lin2024vastgaussian} progressive partitioning for aerial views but faces projection errors with street views during visibility-based camera selection.
%
To better accommodate the reconstruction of large-scale scenes involving both aerial and street views, we make the following adjustments.
Following VastGaussian, 
we first partition the scene into $m \times n$ chunks based on ground-projected camera positions, then expand the original boundaries of each chunk by a threshold to ensure sufficient overlap. Next, we apply different strategies for aerial and street views:
for aerial views, we augment each chunk with additional visible cameras and accessory points,
for street views, which experience significant occlusion, we generate point clouds from depth maps to ensure comprehensive training coverage. All scenes follow the Manhattan world assumption, with the z-axis perpendicular to the ground plane.

Each chunk is trained independently using the strategy discussed in Section~\ref{sec: training_strategy}, then merged into a single model by discarding Gaussians outside the original boundaries and concatenating Gaussian attributes across chunks. To enhance rendering speed, we convert the hybrid representation to a fully explicit one by removing view inputs from each MLP and replacing the MLPs for color with SH predictions.
%

\subsection{Loss Function and Regularization}
\label{sec:loss}
\paragraph{Photometric Loss.}
We use different losses for rendering and surface reconstruction. 
For rendering, in addition to the standard L1, SSIM Loss, we also employ volume regularization $\mathcal{L}_{\mathrm{vol}}$ from Scaffold-GS~\cite{lu2023scaffold} to ensure Gaussian primitives effectively cover the entire scene.

\paragraph{Geometry Regularizations.}
For geometry awareness, we introduce depth supervision $\mathcal{L}_{d}=\frac{1}{h w} \sum \left|\hat{D}-D\right|$, where $h$ and $w$ the rendering image dimensions, $D$ is the inverse of the rendered depth map and $D_s$ is the pseudo ground truth disparity map generated by either rendering engines (for synthetic data) or a scale-aligned pretrained monocular depth model~\cite{yang2024depth} (for real data). For surface reconstruction, we also incorporate the normal loss $\mathcal{L}_n$, as used in 2D-GS~\cite{huang20242d}, to enforce normal consistency.


\paragraph{Mask Regularization.}
To mitigate the effects of moving pedestrians, vehicles, and the sky, which is challenging for vanilla GS methods, we employ an opacity regularization
\begin{equation}
\mathcal{L}_o= -\frac{1}{h w} \sum \sigma \cdot \log \sigma - \frac{1}{h w} \sum M \cdot \log (1-\sigma),
\end{equation}
which encourages the opacity values toward $0$ or $1$, where $M$ denotes the mask region. 
Finally, the combined rendering and surface reconstruction losses, $\mathcal{L}_{R}$ and $\mathcal{L}_{S}$, are defined as follows respectively:
\begin{equation}
    \begin{aligned}
        \mathcal{L}_{R} &= \mathcal{L}_1+\lambda_{\mathrm{ssim}} \mathcal{L}_{\mathrm{ssim}}+\lambda_{\mathrm{vol}} \mathcal{L}_{\mathrm{vol}}+\lambda_d \mathcal{L}_d+\lambda_o \mathcal{L}_o, \\
        \mathcal{L}_{S} &= \mathcal{L}_{R} + \lambda_n \mathcal{L}_n.
    \end{aligned}
\end{equation}
 \section{Dataset Curation}
\label{sec:data}

\begin{table*}[htbp]
\centering
\renewcommand{\arraystretch}{1.15}
\setlength{\tabcolsep}{2pt}
\resizebox{1\linewidth}{!}{
\begin{tabular}{l|ccc|ccc|ccc|ccc|ccc|ccc}
\toprule
 & \multicolumn{6}{c|}{Block\_Small} & \multicolumn{6}{c|}{Synthetic} & \multicolumn{6}{c}{Real} \\ \cmidrule(l){2-19}
\multicolumn{1}{c|}{\multirow{-2}{*}{\centering Scene}} & \multicolumn{3}{c|}{Aerial} & \multicolumn{3}{c|}{Street} & \multicolumn{3}{c|}{Aerial} & \multicolumn{3}{c|}{Street} & \multicolumn{3}{c|}{Aerial} & \multicolumn{3}{c}{Street} \\  \midrule
\begin{tabular}{c|c} Method & Metrics \end{tabular}  & PSNR$\uparrow$ & SSIM$\uparrow$ & LPIPS$\downarrow$ & PSNR$\uparrow$ & SSIM$\uparrow$ & LPIPS$\downarrow$ & PSNR$\uparrow$ & SSIM$\uparrow$ & LPIPS$\downarrow$ & PSNR$\uparrow$ & SSIM$\uparrow$ & LPIPS$\downarrow$ & PSNR$\uparrow$ & SSIM$\uparrow$ & LPIPS$\downarrow$ & PSNR$\uparrow$ & SSIM$\uparrow$ & LPIPS$\downarrow$ \\
\midrule
2D-GS~\cite{huang20242d} & 24.52 & 0.743 & 0.365 & 22.46 & 0.763 & 0.360 & 24.75 & 0.778 & 0.374 & 23.91 & 0.795 & 0.293 & 19.69 & 0.504 & 0.585 & 20.59 & 0.603 & 0.422 \\
Our-2D-GS & \cellcolor{tabsecond}29.60 & \cellcolor{tabsecond}0.899 & \cellcolor{tabsecond}0.121 & \cellcolor{tabthird}23.60 & \cellcolor{tabsecond}0.837 & \cellcolor{tabsecond}0.216 & \cellcolor{tabsecond}30.39 & \cellcolor{tabsecond}0.925 & \cellcolor{tabsecond}0.136 & \cellcolor{tabthird}25.44 & \cellcolor{tabthird}0.848 & \cellcolor{tabthird}0.216 & \cellcolor{tabsecond}22.57 & \cellcolor{tabsecond}0.687 & \cellcolor{tabsecond}0.357 & \cellcolor{tabthird}21.52 & \cellcolor{tabthird}0.655 & \cellcolor{tabthird}0.344 \\
\midrule
3D-GS~\cite{kerbl20233d} & 25.44 & 0.781 & 0.325 & 21.81 & 0.744 & 0.371 & 25.51 & 0.798 & 0.355 & 23.99 & 0.809 & 0.277 & 20.09 & 0.527 & 0.564 & 21.41 & 0.627 & 0.398 \\
Scaffold-GS~\cite{lu2023scaffold} & \cellcolor{tabthird}28.44 & \cellcolor{tabthird}0.863 & 0.191 & \cellcolor{tabfirst}23.84 & \cellcolor{tabthird}0.819 & 0.271 & \cellcolor{tabthird}28.79 & \cellcolor{tabthird}0.891 & \cellcolor{tabthird}0.196 & 25.14 & 0.833 & 0.247 & 20.18 & 0.539 & 0.549 & 21.22 & 0.626 & 0.394 \\
Hier-GS~\cite{kerbl2024hierarchical} & 28.31 & 0.861 & \cellcolor{tabthird}0.189 & 23.75 & 0.823 & \cellcolor{tabthird}0.220 & 28.16 & 0.866 & 0.244 & \cellcolor{tabsecond}25.58 & \cellcolor{tabsecond}0.861 & \cellcolor{tabsecond}0.196 & \cellcolor{tabthird}21.43 & \cellcolor{tabthird}0.639 & \cellcolor{tabthird}0.430 & \cellcolor{tabfirst}22.53 & \cellcolor{tabfirst}0.686 & \cellcolor{tabfirst}0.303 \\
Ours & \cellcolor{tabfirst}30.59 & \cellcolor{tabfirst}0.913 & \cellcolor{tabfirst}0.094 & \cellcolor{tabsecond}23.80 & \cellcolor{tabfirst}0.839 & \cellcolor{tabfirst}0.209 & \cellcolor{tabfirst}31.60 & \cellcolor{tabfirst}0.938 & \cellcolor{tabfirst}0.101 & \cellcolor{tabfirst}25.69 & \cellcolor{tabfirst}0.862 & \cellcolor{tabfirst}0.190 & \cellcolor{tabfirst}23.23 & \cellcolor{tabfirst}0.729 & \cellcolor{tabfirst}0.321 & \cellcolor{tabsecond}22.04 & \cellcolor{tabsecond}0.669 & \cellcolor{tabsecond}0.324 \\
\bottomrule
\end{tabular}}
\vspace{-6pt}
\caption{Quantitative comparison on small-scale datasets. \modelname consistently achieves superior rendering quality compared to baselines in both aerial and street views.}
\label{tab:small_scale}
\vspace{-6pt}
\end{table*}

\begin{figure*}[htbp]
    \centering
    \includegraphics[width=\linewidth]{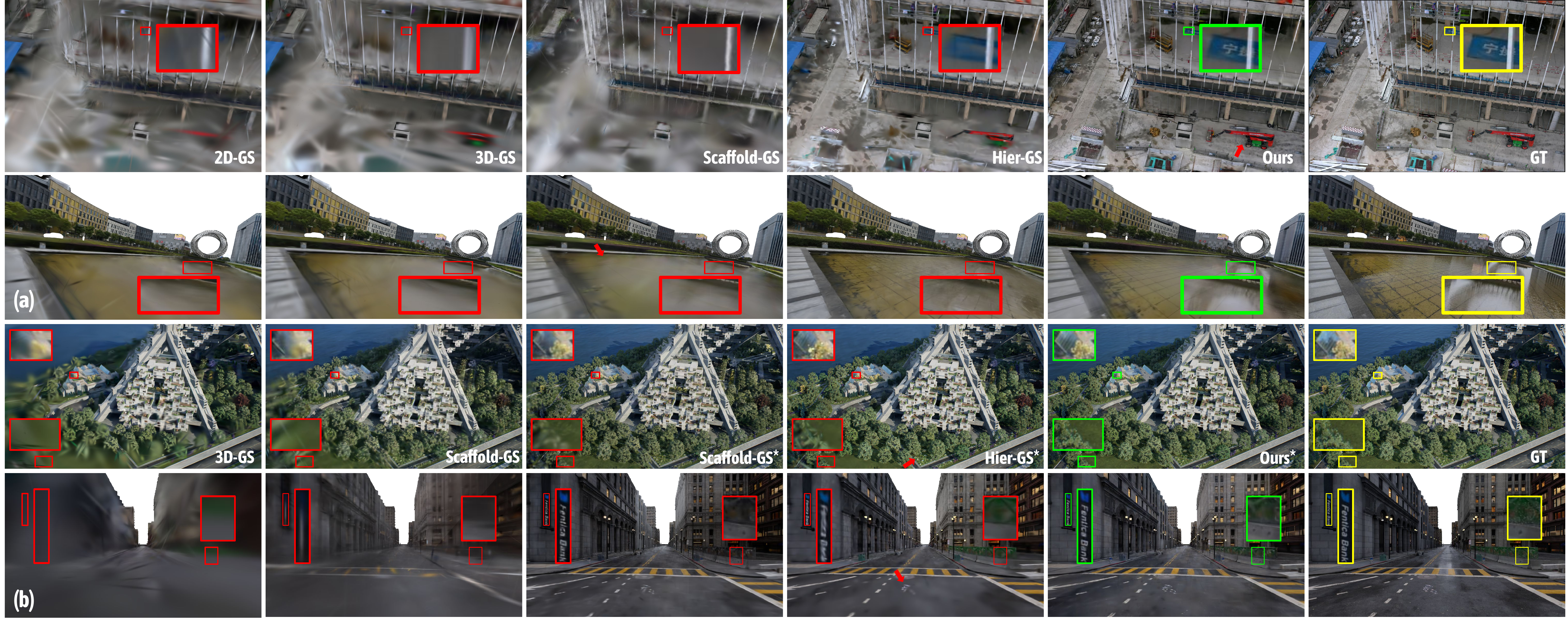}
    \vspace{-0.7cm}
    \caption{Qualiative comparisons of \modelname against baselines~\cite{huang20242d,kerbl20233d,lu2023scaffold,kerbl2024hierarchical} across 
    (a) small-scale and (b) large-scale scenes.}
    \label{fig:main}
    \vspace{-7pt}
\end{figure*}

As previously mentioned, the lack of calibrated aerial and street datasets is a crucial issue that restricts the research on the unified scene reconstruction. To advance the field and evaluate our method, we construct a large-scale dataset with aerial and street views calibrated, which contains five synthetic scenes and two real scenes, as illustrated in Fig.~\ref{fig:data_vis}. 
More details are listed in the supplementary materials.


\paragraph{Synthetic Data.}


Following MatrixCity~\cite{li2023matrixcity} curation process, we create 5 additional synthetic scenes to simulate real-world data using custom data collection plugins. To ensure diversity, these synthetic datasets encompass various scales, seasons, and environments, ranging from realistic cityscapes to imaginative gaming scenarios. In addition to RGB images, we render depth maps for geometry supervision and initial point clouds for 3D-GS training. 
\paragraph{Real Data.} To evaluate \modelname in real-world data, we collect 2 additional scenes. We use a DJI drone equipped with five cameras to capture aerial images and a custom-designed helmet equipped with six Action4 cameras to capture street data. 
To enhance the performance of the structure-from-motion (SfM) technique, we include transitional views to increase image overlap and utilize a powerful commercial software, ContextCapture\footnote{https://www.daspatial.com/cn/gcluster}. Depth maps are generated using Depth-Anything-V2~\cite{yang2024depth}, and scale/offset are estimated by aligning inverse depth of SfM points. Moving objects, such as humans and cars, are removed by Grouned-SAM~\cite{ren2024grounded}, while sky masking is handled by the pretrained SkyRemovel model~\cite{skyremoval}.
More details about the dataset are provided in the supplementary materials.


 \section{Experiments}
\label{sec:exp}



\begin{table*}[htbp]
\centering
\renewcommand{\arraystretch}{1.15}
\setlength{\tabcolsep}{2pt}
\resizebox{1\linewidth}{!}{
\begin{tabular}{l|ccc|ccc|ccc|ccc|ccc|ccc}
\toprule
 & \multicolumn{6}{c|}{Block\_A} & \multicolumn{6}{c|}{Hillside\_Morning} & \multicolumn{6}{c}{Hillside\_Summer} \\ \cmidrule(l){2-19}
\multicolumn{1}{c|}{\multirow{-2}{*}{\centering Scene}} & \multicolumn{3}{c|}{Aerial} & \multicolumn{3}{c|}{Street} & \multicolumn{3}{c|}{Aerial} & \multicolumn{3}{c|}{Street} & \multicolumn{3}{c|}{Aerial} & \multicolumn{3}{c}{Street} \\  \midrule
\begin{tabular}{c|c} Method & Metrics \end{tabular} & PSNR$\uparrow$ & SSIM$\uparrow$ & LPIPS$\downarrow$ & PSNR$\uparrow$ & SSIM$\uparrow$ & LPIPS$\downarrow$ & PSNR$\uparrow$ & SSIM$\uparrow$ & LPIPS$\downarrow$ & PSNR$\uparrow$ & SSIM$\uparrow$ & LPIPS$\downarrow$ & PSNR$\uparrow$ & SSIM$\uparrow$ & LPIPS$\downarrow$ & PSNR$\uparrow$ & SSIM$\uparrow$ & LPIPS$\downarrow$ \\
\midrule

2D-GS~\cite{huang20242d} & 20.63 & 0.585 & 0.595 & 19.57 & 0.635 & 0.477 & 22.02 & 0.629 & 0.513 & 19.20 & 0.637 & 0.470 & 20.87 & 0.560 & 0.524 & 18.70 & 0.614 & 0.481 \\

Our-2D-GS* & \cellcolor{tabsecond}28.52 & \cellcolor{tabsecond}0.886 & \cellcolor{tabsecond}0.166 & 23.00 & 0.789 & 0.301 & \cellcolor{tabsecond}29.98 & \cellcolor{tabsecond}0.900 & \cellcolor{tabsecond}0.135 & 20.44 & \cellcolor{tabthird}0.686 & \cellcolor{tabthird}0.404 & \cellcolor{tabsecond}28.88 & \cellcolor{tabsecond}0.889 & \cellcolor{tabsecond}0.126 & 20.23 & \cellcolor{tabthird}0.674 & \cellcolor{tabthird}0.404 \\

\midrule

3D-GS~\cite{kerbl20233d} & 21.02 & 0.590 & 0.595 & 19.06 & 0.628 & 0.477 & 
22.87 & 0.662 & 0.489 & 19.35 & 0.643 & 0.466 & 21.50 & 0.592 & 0.503 & 18.63 & 0.617 & 0.480 \\

Scaffold-GS~\cite{lu2023scaffold} & 22.98 & 0.654 & 0.523 & 20.89 & 0.681 & 0.436 & 
23.05 & 0.688 & 0.433 & 20.08 & 0.656 & 0.461 & 22.02 & 0.634 & 0.436 & 19.36 & 0.629 & 0.473 \\

Scaffold-GS* & \cellcolor{tabthird}27.62 & \cellcolor{tabthird}0.860 & \cellcolor{tabthird}0.206 & \cellcolor{tabthird}23.10 & \cellcolor{tabthird}0.808 & \cellcolor{tabthird}0.277 & \cellcolor{tabthird}25.46 & \cellcolor{tabthird}0.796 & 0.271 & \cellcolor{tabsecond}20.66 & 0.685 & 0.416 & \cellcolor{tabthird}23.91 & \cellcolor{tabthird}0.752 & 0.279 & \cellcolor{tabthird}20.30 & 0.667 & 0.423 \\

Hier-GS*~\cite{kerbl2024hierarchical} &  26.42 & 0.792 & 0.340 & \cellcolor{tabsecond}23.59 & \cellcolor{tabsecond}0.811 & \cellcolor{tabsecond}0.275 & 25.41 & 0.783 & \cellcolor{tabthird}0.269 & \cellcolor{tabfirst}21.13 & \cellcolor{tabfirst}0.692 & \cellcolor{tabfirst}0.398 & 23.53 & 0.735 & \cellcolor{tabthird}0.273 & \cellcolor{tabfirst}20.47 & \cellcolor{tabsecond}0.673 & \cellcolor{tabsecond}0.402\\ 

Ours* & \cellcolor{tabfirst}28.89 & \cellcolor{tabfirst}0.890 & \cellcolor{tabfirst}0.151 & \cellcolor{tabfirst}23.66 & \cellcolor{tabfirst}0.816 & \cellcolor{tabfirst}0.255 & \cellcolor{tabfirst}31.09 & \cellcolor{tabfirst}0.918 & \cellcolor{tabfirst}0.107 & \cellcolor{tabthird}20.62 & \cellcolor{tabsecond}0.689 & \cellcolor{tabsecond}0.399 & \cellcolor{tabfirst}29.34 & \cellcolor{tabfirst}0.900 & \cellcolor{tabfirst}0.110 & \cellcolor{tabsecond}20.34 & \cellcolor{tabfirst}0.681 & \cellcolor{tabfirst}0.393\\

\bottomrule
\end{tabular}}
\vspace{-6pt}
\caption{Quantitative comparison on large-scale datasets. \modelname achieves superior rendering quality compared to baselines. }
\label{tab:large_scale}
\vspace{-6pt}
\end{table*}

\subsection{Experimental Setup}
\paragraph{Datasets and Metrics.}
We conduct comprehensive evaluations across 11 scenes containing both aerial and street views, sourced from the MatrixCity dataset~\cite{li2023matrixcity}, the UC-GS dataset~\cite{zhang2024drone}, and our captured dataset.
Specifically, in the MatrixCity dataset, we evaluate two synthetic scenes: Block\_Small and Block\_A. 
For the UC-GS dataset, we select two synthetic scenes, New York City (NYC) and San Francisco (SF). Consistent with the original UC-GS settings, evaluations are performed solely on street views. 
For our newly captured dataset, we select one out of every 32 images for evaluation.
Among them, we divide Block\_A into $4 \times 2$ chunks and divide Hillside\_Morning and Hillside\_Summer into $6 \times 2$ chunks to improve GPU compatibility and achieve satisfactory visual results.
We use image resolution 1920$\times$1080 for synthetic data; 1600$\times$1066 and 1600$\times$900 for real-world aerial and street view images.

Results are evaluated using standard visual quality metrics: PSNR, SSIM~\cite{wang2004image}, and LPIPS~\cite{zhang2018unreasonable}. Aerial and street view images are evaluated separately. To minimize the impact of pedestrians, cars, and the sky, we mask out those moving entities out when comparing with ground truths.

\paragraph{Baselines.}
We compare our method against 3D-GS~\cite{kerbl20233d}, Scaffold-GS~\cite{lu2023scaffold}, UC-GS~\cite{zhang2024drone} and Hierarchical-3DGS~\cite{kerbl2024hierarchical} for rendering, as well as 2D-GS~\cite{huang20242d} for surface reconstruction. To ensure full convergence, we extend the total training iterations from 30k to 100k, with densification continuing until the 50k iteration. For all baselines, we run their official code on all datasets except for the UC-GS dataset, where we adopt the metrics from their original paper to maintain consistency with their settings.  For Hierarchical-3DGS, we report the best visual quality results after optimizing the hierarchy (leaves). For each baseline, we also add depth supervision same as our method. 
In the large-scale setting, we partition the scenes and train each chunk in parallel, using $^*$ to denote the strategy discussed in Sec.~\ref{sec:urban-strategy}. 

\paragraph{Implementation Details.}
For a fair comparison, we train the first stage for 60k iterations and the second stage for 40k iterations, with densification stopping at 30k and 20k iterations, respectively. In the first stage, we follow the default settings of Scaffold-GS~\cite{lu2023scaffold} for learning rate and densification. In the second stage, we reduce the learning rate of the offset by a factor of 10 and set $\tau_\sigma$, $\tau_g$ and $N$ to 0.2, 0.15 and 100, respectively. In LOD construction stage, $r_d$ is set to 0.999. The loss function parameters are $\lambda_\mathrm{ssim} = 0.2$ , $\lambda_\mathrm{vol} = 0.01$, $\lambda_o = 0.05$, and $\lambda_n = 0.05$. The weight for $\mathcal{L}_d$ is exponentially decayed from 1 to 0.01 over both stages. Depth supervision is activated after 500 iterations, and normal supervision after 7k iterations. We denote neural Gaussians that MLP decode into 2D as `Our-2D-GS' and denote `Ours' if the primitives are generated into 3D.  
All experiments are conducted on NVIDIA A100 80G GPUs. 

\subsection{Results Analysis}
\begin{figure}[t!]
\centering
    \includegraphics[width=\linewidth]{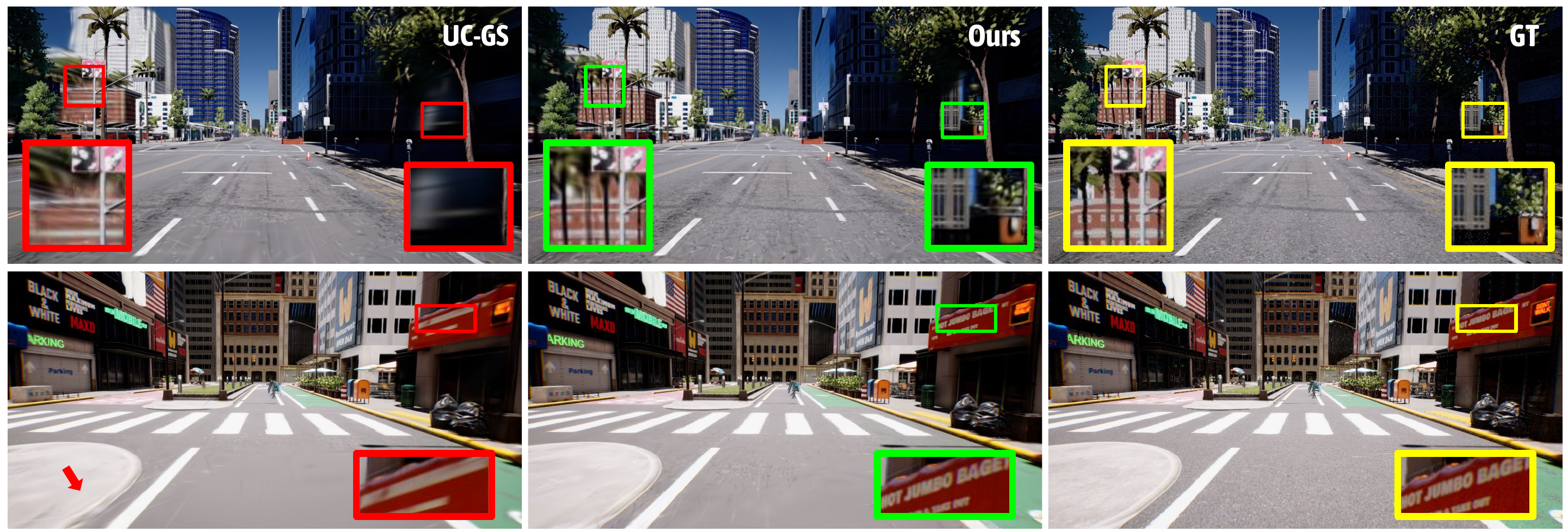}
    \vspace{-0.7cm}
    \caption{Qualitative comparisons of \modelname against UC-GS~\cite{zhang2024drone}. The first row shows view shifting in the SF scene, while the second row shows view shifting and rotation in the NYC scene.}
    \vspace{-7pt}
    \label{fig:ucgs}
\end{figure}

\begin{figure*}[htbp]
    \centering
    \includegraphics[width=\linewidth]{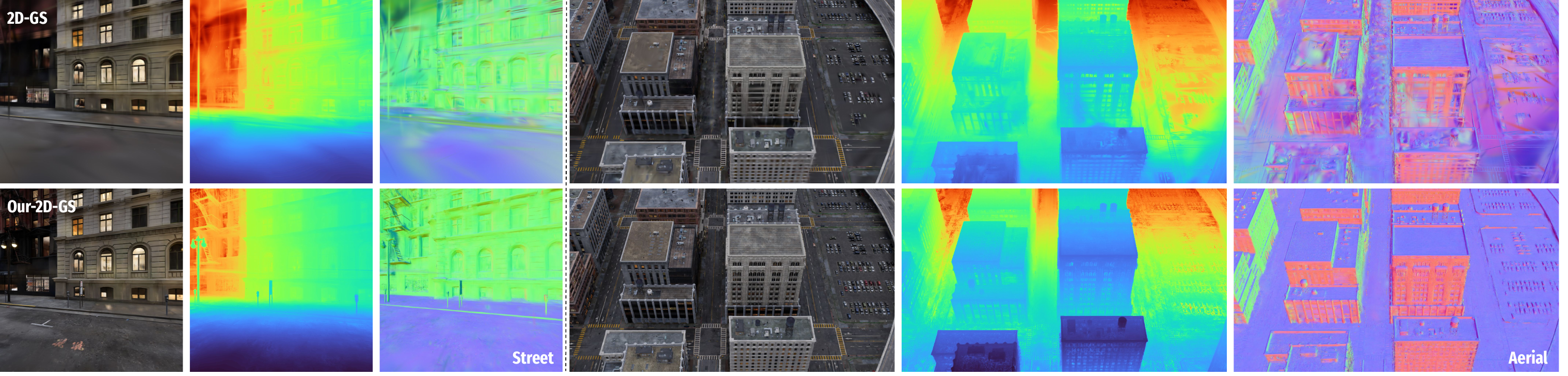}
    \vspace{-0.7cm}
    \caption{Qualitative comparisons of our method with 2D-GS~\cite{huang20242d}. We visualize the rendered images and geometry results (depth and world-space normal) in the Block\_Small scene from both aerial and street views.}
    \vspace{-7pt}
    \label{fig:surface}
\end{figure*}

Our evaluation covers diverse scenes, including aerial and street views, intricate and large-scale scenes, both synthetic and real-world. We demonstrate that \modelname preserves fine details and significantly improves the quality of surface reconstruction , as shown in Tab.~\ref{tab:small_scale},~\ref{tab:ucgs},~\ref{tab:large_scale} and Fig.~\ref{fig:main},~\ref{fig:ucgs},~\ref{fig:surface}.

\begin{table}[t!]
\centering
\renewcommand{\arraystretch}{1.15}
\setlength{\tabcolsep}{1pt}
\resizebox{1\linewidth}{!}{
\begin{tabular}{l|ccc|ccc|ccc}
\toprule
 \multicolumn{1}{c|}{\centering Scene} & \multicolumn{3}{c|}{Held-out} & \multicolumn{3}{c|}{View($+1$m)} & \multicolumn{3}{c}{View($+1$m $5^{\circ}$down)}  \\  
\begin{tabular}{c|c} Method & Metrics \end{tabular}  & PSNR$\uparrow$ & SSIM$\uparrow$ & LPIPS$\downarrow$ & PSNR$\uparrow$ & SSIM$\uparrow$ & LPIPS$\downarrow$ & PSNR$\uparrow$ & SSIM$\uparrow$ & LPIPS$\downarrow$  \\
\midrule

3D-GS~\cite{kerbl20233d} & 23.47 & 0.668 & 0.406 & 20.83 & 0.605 & 0.440 & 21.25 & 0.643 & 0.402\\

Scaffold-GS~\cite{lu2023scaffold} &  25.40 & 0.744 & 0.320 & 22.62 & 0.671 & 0.375 & 23.28 & 0.711 & 0.334\\

UC-GS~\cite{lu2023scaffold} & \textbf{25.95} & \textbf{0.763} & \underline{0.291} & \underline{23.52} & \underline{0.702} & \underline{0.340} & \underline{24.15} & \underline{0.741} & \underline{0.298} \\

Ours & \underline{25.35} & \underline{0.757} & \textbf{0.280} & \textbf{25.46} & \textbf{0.760} & \textbf{0.280} & \textbf{25.37} & \textbf{0.761} & \textbf{0.278}\\

\bottomrule
\end{tabular}}
\vspace{-6pt}
\caption{Quantitative comparison on UC-GS dataset~\cite{zhang2024drone}. The metrics are extracted from the original paper, in which all methods are trained for 900k iterations.}
\label{tab:ucgs}
\vspace{-6pt}
\end{table}

\paragraph{Rendering Performance Analysis.} Our method reconstructs a unified Gaussian model, delivering excellent visual quality in both aerial and street views. Compared to 2D-GS~\cite{huang20242d}, Our-2D-GS consistently outperforms by 4 dB in aerial PSNR and 1.5 dB in street PSNR. Furthermore, compared to all the baselines, \modelname exceeds the performance of all baselines, except Hierarchical-3DGS, which is competitive in street views due to its street-focused design.
As shown in the highlighted patches and arrows above of Fig.~\ref{fig:main} (a), our method excels in fine details (1st row) and reflective regions (2nd row).
In the UC-GS setting (Tab.~\ref{tab:ucgs}), under the held-out viewpoints condition where test views share the same heights as training views, \modelname, trained for 100k iterations, shows slightly lower PSNR and SSIM values compared to UC-GS trained for 900k iterations. However, our approach produces richer details, as indicated by better LPIPS metrics. Moreover, when evaluating shifted and rotated views, our method consistently outperforms the baselines, as shown in Fig.~\ref{fig:ucgs}.

\paragraph{Large-scale Rendering Performance.}
\label{sec:large_scale_result}
For the challenging large-scale scenes, we conduct two types of experiments: direct training without chunking and chunked training.
As shown in Table~\ref{tab:large_scale}, partitioning and then merging proves effective, especially when comparing Scaffold-GS and Scaffold-GS*. \modelname consistently delivers excellent rendering quality in large-scale scenes compared to other baselines. 
In contrast, Hierarchical-3DGS~\cite{kerbl2024hierarchical} tends to overfit the training views, producing artifacts in novel views, while other baselines fail to capture fine-grained details (Figure~\ref{fig:main} (b)). 
Each scene is divided into chunks and trained in parallel, with a total training time of about four hours per scene.
Due to the LOD design, our method achieves real-time rendering speeds of 51.5 FPS in the Hillside\_Morning and Hillside\_Summer scenes, which are the largest scenes among our captured dataset.

\paragraph{Surface Reconstruction Performance.} As shown in Fig.\ref{fig:surface}, our method reconstructs more complete and detailed geometry compared to 2D-GS~\cite{huang20242d}. 2D-GS produces artifacts, resulting in incomplete and lackluster geometry. In contrast, our method, using the two-stage training approach and enhanced densification, delivers detailed, geometrically accurate, and artifact-free reconstruction.


\subsection{Ablation Studies}
\label{sec:ablation}
In this section, we ablate each individual module to validate their effectiveness. We use the scenes from our proposed real dataset, Road and Park. Quantitative results can be found in Tab.~\ref{tab:ablate}.

\paragraph{Balanced Camera Distribution.} To assess the effectiveness of the balanced camera distribution (Sec.~\ref{sec: camera_balance}), we conduct an ablation study with randomly selected training views. The results in Tab.~\ref{tab:ablate} reveal a notable decline in the visual quality of aerial views, while the quality of street views remains similar. This suggests that a well-distributed set of observation perspectives enhances the model performance, rather than converging to a single-level view.



\paragraph{Balanced Densification.} As outlined in Sec.~\ref{sec: training_strategy}, we meticulously design the densification strategy in the two training stages. Specifically, gradients of the Gaussian primitives are accumulated from aerial views, while in the second stage, they are derived solely from street views. To assess the effectiveness of this approach, we compute the gradients of the Gaussian primitives across all the images in both stages. The results reveal conflicts during the densification process, highlighting the need to explicitly separate the densification steps during optimization.

\begin{table}[t!]
\centering
\renewcommand{\arraystretch}{1.1}
\setlength{\tabcolsep}{3pt}
\resizebox{1\linewidth}{!}{
\begin{tabular}{l|ccc|ccc}
\toprule
\multicolumn{1}{c|}{\centering Scene} & \multicolumn{3}{c|}{Aerial} & \multicolumn{3}{c}{Street} \\  
\begin{tabular}{c|c} Method & Metrics \end{tabular}  & PSNR$\uparrow$ & SSIM$\uparrow$ & LPIPS$\downarrow$ & PSNR$\uparrow$ & SSIM$\uparrow$ & LPIPS$\downarrow$  \\
\midrule


Ours w/o camera bal. & 22.49 & 0.684 & 0.375 & 22.00 & \underline{0.671} & \underline{0.323} \\

Ours w/o densify bal. & 23.02 & \underline{0.721} & \underline{0.332} & 21.98 & 0.670 & 0.325 \\

Ours w/o multi LOD & \underline{23.05} & 0.717 & 0.337 & \underline{22.03} & \textbf{0.672} & \textbf{0.321} \\


Ours w/o densify poli. & 22.83 & 0.690 & 0.369 & 21.68 & 0.655 & 0.346 \\

Ours  & \textbf{23.23} & \textbf{0.729} & \textbf{0.321} & \textbf{22.04} & 0.669 & 0.324 \\

\bottomrule
\end{tabular}}
\vspace{-6pt}
\caption{Ablations of our method on our proposed real datasets.}
\label{tab:ablate}
\vspace{-1em}
\end{table}

\paragraph{Multi-Resolution LOD.} To assess the impact of LOD design, we conduct an experiment without LOD. The quality of aerial views significantly decreases, indicating that the strategy enhances the capture of fine-scale details. 



\paragraph{Densification Policy.} We perform an ablation of the densification policy (Eq.~\ref{eq:den_policy}) by replacing it in the second stage with the one from Scaffold-GS. This leads to a significant reduction in rendering details for both aerial and street views, indicating that relying solely on average gradients for spawning Gaussian primitives is insufficient to capture local fine-grained details.


 \section{Conclusion}
\label{sec:conclusion}
We introduce \modelname, a novel framework for large-scale aerial-to-ground scene reconstruction. We first explore the challenges of unified scene reconstruction from both aerial and street views, and propose a end-to-end coarse-to-fine training framework that mitigates inherent conflicts and is adaptable for large-scale reconstruction. Additionally, we present a comprehensive dataset with two cross views. Our experiments show that \modelname outperforms existing methods in visual quality and geometry accuracy. 
In the future, we aim to minimize input dependency by leveraging advanced techniques such as pre-trained 3D scene foundation models and enable more systematic solutions, such as distributional training.

{
    \small
    \bibliographystyle{ieeenat_fullname}
    \bibliography{main}

\begin{thebibliography}{47}
\providecommand{\natexlab}[1]{#1}
\providecommand{\url}[1]{\texttt{#1}}
\expandafter\ifx\csname urlstyle\endcsname\relax
  \providecommand{\doi}[1]{doi: #1}\else
  \providecommand{\doi}{doi: \begingroup \urlstyle{rm}\Url}\fi

\bibitem[Bojanowski et~al.(2017)Bojanowski, Joulin, Lopez-Paz, and Szlam]{bojanowski2017optimizing}
Piotr Bojanowski, Armand Joulin, David Lopez-Paz, and Arthur Szlam.
\newblock Optimizing the latent space of generative networks.
\newblock \emph{arXiv preprint arXiv:1707.05776}, 2017.

\bibitem[Chen et~al.(2022)Chen, Xu, Geiger, Yu, and Su]{chen2022tensorf}
Anpei Chen, Zexiang Xu, Andreas Geiger, Jingyi Yu, and Hao Su.
\newblock Tensorf: Tensorial radiance fields.
\newblock In \emph{European Conference on Computer Vision}, pages 333--350. Springer, 2022.

\bibitem[Chen and Lee(2024)]{chen2024dogaussian}
Yu Chen and Gim~Hee Lee.
\newblock Dogaussian: Distributed-oriented gaussian splatting for large-scale 3d reconstruction via gaussian consensus.
\newblock \emph{arXiv preprint arXiv:2405.13943}, 2024.

\bibitem[Chen et~al.(2023)Chen, Gu, Jiang, Zhu, and Zhang]{chen2023periodic}
Yurui Chen, Chun Gu, Junzhe Jiang, Xiatian Zhu, and Li Zhang.
\newblock Periodic vibration gaussian: Dynamic urban scene reconstruction and real-time rendering.
\newblock \emph{arXiv:2311.18561}, 2023.

\bibitem[Crandall et~al.(2012)Crandall, Owens, Snavely, and Huttenlocher]{crandall2012sfm}
David~J Crandall, Andrew Owens, Noah Snavely, and Daniel~P Huttenlocher.
\newblock Sfm with mrfs: Discrete-continuous optimization for large-scale structure from motion.
\newblock \emph{IEEE transactions on pattern analysis and machine intelligence}, 35\penalty0 (12):\penalty0 2841--2853, 2012.

\bibitem[Fridovich-Keil et~al.(2022)Fridovich-Keil, Yu, Tancik, Chen, Recht, and Kanazawa]{fridovich2022plenoxels}
Sara Fridovich-Keil, Alex Yu, Matthew Tancik, Qinhong Chen, Benjamin Recht, and Angjoo Kanazawa.
\newblock Plenoxels: Radiance fields without neural networks.
\newblock In \emph{Proceedings of the IEEE/CVF Conference on Computer Vision and Pattern Recognition}, pages 5501--5510, 2022.

\bibitem[Gu{\'e}don and Lepetit(2024)]{guedon2024sugar}
Antoine Gu{\'e}don and Vincent Lepetit.
\newblock Sugar: Surface-aligned gaussian splatting for efficient 3d mesh reconstruction and high-quality mesh rendering.
\newblock In \emph{Proceedings of the IEEE/CVF Conference on Computer Vision and Pattern Recognition}, pages 5354--5363, 2024.

\bibitem[Huang et~al.(2024)Huang, Yu, Chen, Geiger, and Gao]{huang20242d}
Binbin Huang, Zehao Yu, Anpei Chen, Andreas Geiger, and Shenghua Gao.
\newblock 2d gaussian splatting for geometrically accurate radiance fields.
\newblock In \emph{ACM SIGGRAPH 2024 Conference Papers}, pages 1--11, 2024.

\bibitem[Kerbl et~al.(2023)Kerbl, Kopanas, Leimk{\"u}hler, and Drettakis]{kerbl20233d}
Bernhard Kerbl, Georgios Kopanas, Thomas Leimk{\"u}hler, and George Drettakis.
\newblock 3d gaussian splatting for real-time radiance field rendering.
\newblock \emph{ACM Transactions on Graphics}, 42\penalty0 (4), 2023.

\bibitem[Kerbl et~al.(2024)Kerbl, Meuleman, Kopanas, Wimmer, Lanvin, and Drettakis]{kerbl2024hierarchical}
Bernhard Kerbl, Andreas Meuleman, Georgios Kopanas, Michael Wimmer, Alexandre Lanvin, and George Drettakis.
\newblock A hierarchical 3d gaussian representation for real-time rendering of very large datasets.
\newblock \emph{ACM Transactions on Graphics (TOG)}, 43\penalty0 (4):\penalty0 1--15, 2024.

\bibitem[Li et~al.(2024)Li, Chen, Wang, Liao, Yan, and Xiong]{li2024retinags}
Bingling Li, Shengyi Chen, Luchao Wang, Kaimin Liao, Sijie Yan, and Yuanjun Xiong.
\newblock Retinags: Scalable training for dense scene rendering with billion-scale 3d gaussians.
\newblock \emph{arXiv preprint arXiv:2406.11836}, 2024.

\bibitem[Li et~al.(2023)Li, Jiang, Xu, Xiangli, Wang, Lin, and Dai]{li2023matrixcity}
Yixuan Li, Lihan Jiang, Linning Xu, Yuanbo Xiangli, Zhenzhi Wang, Dahua Lin, and Bo Dai.
\newblock Matrixcity: A large-scale city dataset for city-scale neural rendering and beyond.
\newblock In \emph{Proceedings of the IEEE/CVF International Conference on Computer Vision}, pages 3205--3215, 2023.

\bibitem[Liao et~al.(2022)Liao, Xie, and Geiger]{liao2022kitti}
Yiyi Liao, Jun Xie, and Andreas Geiger.
\newblock Kitti-360: A novel dataset and benchmarks for urban scene understanding in 2d and 3d.
\newblock \emph{IEEE Transactions on Pattern Analysis and Machine Intelligence}, 45\penalty0 (3):\penalty0 3292--3310, 2022.

\bibitem[Lin et~al.(2024)Lin, Li, Tang, Liu, Liu, Liu, Lu, Wu, Xu, Yan, et~al.]{lin2024vastgaussian}
Jiaqi Lin, Zhihao Li, Xiao Tang, Jianzhuang Liu, Shiyong Liu, Jiayue Liu, Yangdi Lu, Xiaofei Wu, Songcen Xu, Youliang Yan, et~al.
\newblock Vastgaussian: Vast 3d gaussians for large scene reconstruction.
\newblock In \emph{Proceedings of the IEEE/CVF Conference on Computer Vision and Pattern Recognition}, pages 5166--5175, 2024.

\bibitem[Lin et~al.(2022)Lin, Liu, Hu, Yan, Xie, and Huang]{lin2022capturing}
Liqiang Lin, Yilin Liu, Yue Hu, Xingguang Yan, Ke Xie, and Hui Huang.
\newblock Capturing, reconstructing, and simulating: the urbanscene3d dataset.
\newblock In \emph{European Conference on Computer Vision}, pages 93--109. Springer, 2022.

\bibitem[Liu et~al.(2023)Liu, Chen, Yang, Wang, Manivasagam, and Urtasun]{liu2023real}
Jeffrey~Yunfan Liu, Yun Chen, Ze Yang, Jingkang Wang, Sivabalan Manivasagam, and Raquel Urtasun.
\newblock Real-time neural rasterization for large scenes.
\newblock In \emph{Proceedings of the IEEE/CVF International Conference on Computer Vision}, pages 8416--8427, 2023.

\bibitem[Liu et~al.(2020)Liu, Gu, Zaw~Lin, Chua, and Theobalt]{liu2020neural}
Lingjie Liu, Jiatao Gu, Kyaw Zaw~Lin, Tat-Seng Chua, and Christian Theobalt.
\newblock Neural sparse voxel fields.
\newblock \emph{Advances in Neural Information Processing Systems}, 33:\penalty0 15651--15663, 2020.

\bibitem[Liu et~al.(2024)Liu, Guan, Luo, Fan, Peng, and Zhang]{liu2024citygaussian}
Yang Liu, He Guan, Chuanchen Luo, Lue Fan, Junran Peng, and Zhaoxiang Zhang.
\newblock Citygaussian: Real-time high-quality large-scale scene rendering with gaussians.
\newblock \emph{arXiv preprint arXiv:2404.01133}, 2024.

\bibitem[Lu et~al.(2023)Lu, Yu, Xu, Xiangli, Wang, Lin, and Dai]{lu2023scaffold}
Tao Lu, Mulin Yu, Linning Xu, Yuanbo Xiangli, Limin Wang, Dahua Lin, and Bo Dai.
\newblock Scaffold-gs: Structured 3d gaussians for view-adaptive rendering.
\newblock \emph{arXiv preprint arXiv:2312.00109}, 2023.

\bibitem[Mildenhall et~al.(2021)Mildenhall, Srinivasan, Tancik, Barron, Ramamoorthi, and Ng]{mildenhall2021nerf}
Ben Mildenhall, Pratul~P Srinivasan, Matthew Tancik, Jonathan~T Barron, Ravi Ramamoorthi, and Ren Ng.
\newblock Nerf: Representing scenes as neural radiance fields for view synthesis.
\newblock \emph{Communications of the ACM}, 65\penalty0 (1):\penalty0 99--106, 2021.

\bibitem[M{\"u}ller et~al.(2022)M{\"u}ller, Evans, Schied, and Keller]{muller2022instant}
Thomas M{\"u}ller, Alex Evans, Christoph Schied, and Alexander Keller.
\newblock Instant neural graphics primitives with a multiresolution hash encoding.
\newblock \emph{ACM Transactions on Graphics (ToG)}, 41\penalty0 (4):\penalty0 1--15, 2022.

\bibitem[OpenDroneMap(2022)]{skyremoval}
OpenDroneMap.
\newblock Skyremoval.
\newblock \url{https://github.com/OpenDroneMap/SkyRemoval/}, 2022.

\bibitem[Rematas et~al.(2022)Rematas, Liu, Srinivasan, Barron, Tagliasacchi, Funkhouser, and Ferrari]{rematas2022urban}
Konstantinos Rematas, Andrew Liu, Pratul~P Srinivasan, Jonathan~T Barron, Andrea Tagliasacchi, Thomas Funkhouser, and Vittorio Ferrari.
\newblock Urban radiance fields.
\newblock In \emph{Proceedings of the IEEE/CVF Conference on Computer Vision and Pattern Recognition}, pages 12932--12942, 2022.

\bibitem[Ren et~al.(2024{\natexlab{a}})Ren, Jiang, Lu, Yu, Xu, Ni, and Dai]{ren2024octree}
Kerui Ren, Lihan Jiang, Tao Lu, Mulin Yu, Linning Xu, Zhangkai Ni, and Bo Dai.
\newblock Octree-gs: Towards consistent real-time rendering with lod-structured 3d gaussians.
\newblock \emph{arXiv preprint arXiv:2403.17898}, 2024{\natexlab{a}}.

\bibitem[Ren et~al.(2024{\natexlab{b}})Ren, Liu, Zeng, Lin, Li, Cao, Chen, Huang, Chen, Yan, Zeng, Zhang, Li, Yang, Li, Jiang, and Zhang]{ren2024grounded}
Tianhe Ren, Shilong Liu, Ailing Zeng, Jing Lin, Kunchang Li, He Cao, Jiayu Chen, Xinyu Huang, Yukang Chen, Feng Yan, Zhaoyang Zeng, Hao Zhang, Feng Li, Jie Yang, Hongyang Li, Qing Jiang, and Lei Zhang.
\newblock Grounded sam: Assembling open-world models for diverse visual tasks, 2024{\natexlab{b}}.

\bibitem[Sun et~al.(2022)Sun, Sun, and Chen]{sun2022direct}
Cheng Sun, Min Sun, and Hwann-Tzong Chen.
\newblock Direct voxel grid optimization: Super-fast convergence for radiance fields reconstruction.
\newblock In \emph{Proceedings of the IEEE/CVF Conference on Computer Vision and Pattern Recognition}, pages 5459--5469, 2022.

\bibitem[Tancik et~al.(2022)Tancik, Casser, Yan, Pradhan, Mildenhall, Srinivasan, Barron, and Kretzschmar]{tancik2022block}
Matthew Tancik, Vincent Casser, Xinchen Yan, Sabeek Pradhan, Ben Mildenhall, Pratul~P Srinivasan, Jonathan~T Barron, and Henrik Kretzschmar.
\newblock Block-nerf: Scalable large scene neural view synthesis.
\newblock In \emph{Proceedings of the IEEE/CVF Conference on Computer Vision and Pattern Recognition}, pages 8248--8258, 2022.

\bibitem[Tang et~al.(2023)Tang, Ren, Zhou, Liu, and Zeng]{tang2023dreamgaussian}
Jiaxiang Tang, Jiawei Ren, Hang Zhou, Ziwei Liu, and Gang Zeng.
\newblock Dreamgaussian: Generative gaussian splatting for efficient 3d content creation.
\newblock \emph{arXiv preprint arXiv:2309.16653}, 2023.

\bibitem[Tang et~al.(2024)Tang, Chen, Chen, Wang, Zeng, and Liu]{tang2024lgm}
Jiaxiang Tang, Zhaoxi Chen, Xiaokang Chen, Tengfei Wang, Gang Zeng, and Ziwei Liu.
\newblock Lgm: Large multi-view gaussian model for high-resolution 3d content creation.
\newblock \emph{arXiv preprint arXiv:2402.05054}, 2024.

\bibitem[Turki et~al.(2022)Turki, Ramanan, and Satyanarayanan]{turki2022mega}
Haithem Turki, Deva Ramanan, and Mahadev Satyanarayanan.
\newblock Mega-nerf: Scalable construction of large-scale nerfs for virtual fly-throughs.
\newblock In \emph{Proceedings of the IEEE/CVF Conference on Computer Vision and Pattern Recognition}, pages 12922--12931, 2022.

\bibitem[Wang et~al.(2004)Wang, Bovik, Sheikh, and Simoncelli]{wang2004image}
Zhou Wang, Alan~C Bovik, Hamid~R Sheikh, and Eero~P Simoncelli.
\newblock Image quality assessment: from error visibility to structural similarity.
\newblock \emph{IEEE transactions on image processing}, 13\penalty0 (4):\penalty0 600--612, 2004.

\bibitem[Wei et~al.(2024)Wei, Wang, Lu, Xu, Liu, Zhao, Chen, and Wang]{wei2024editable}
Yuxi Wei, Zi Wang, Yifan Lu, Chenxin Xu, Changxing Liu, Hao Zhao, Siheng Chen, and Yanfeng Wang.
\newblock Editable scene simulation for autonomous driving via collaborative llm-agents.
\newblock In \emph{Proceedings of the IEEE/CVF Conference on Computer Vision and Pattern Recognition}, pages 15077--15087, 2024.

\bibitem[Xiangli et~al.(2022)Xiangli, Xu, Pan, Zhao, Rao, Theobalt, Dai, and Lin]{xiangli2022bungeenerf}
Yuanbo Xiangli, Linning Xu, Xingang Pan, Nanxuan Zhao, Anyi Rao, Christian Theobalt, Bo Dai, and Dahua Lin.
\newblock Bungeenerf: Progressive neural radiance field for extreme multi-scale scene rendering.
\newblock In \emph{European conference on computer vision}, pages 106--122. Springer, 2022.

\bibitem[Xiangli et~al.(2023)Xiangli, Xu, Pan, Zhao, Dai, and Lin]{xiangli2023assetfield}
Yuanbo Xiangli, Linning Xu, Xingang Pan, Nanxuan Zhao, Bo Dai, and Dahua Lin.
\newblock Assetfield: Assets mining and reconfiguration in ground feature plane representation.
\newblock \emph{arXiv preprint arXiv:2303.13953}, 2023.

\bibitem[Xiao et~al.(2024)Xiao, Cruz, Ahmedt-Aristizabal, Salvado, Fookes, and Lebrat]{xiao2024nerf}
Wenhui Xiao, Rodrigo~Santa Cruz, David Ahmedt-Aristizabal, Olivier Salvado, Clinton Fookes, and Leo Lebrat.
\newblock Nerf director: Revisiting view selection in neural volume rendering.
\newblock In \emph{Proceedings of the IEEE/CVF Conference on Computer Vision and Pattern Recognition}, pages 20742--20751, 2024.

\bibitem[Xu et~al.(2023)Xu, Xiangli, Peng, Pan, Zhao, Theobalt, Dai, and Lin]{xu2023grid}
Linning Xu, Yuanbo Xiangli, Sida Peng, Xingang Pan, Nanxuan Zhao, Christian Theobalt, Bo Dai, and Dahua Lin.
\newblock Grid-guided neural radiance fields for large urban scenes.
\newblock In \emph{Proceedings of the IEEE/CVF Conference on Computer Vision and Pattern Recognition}, pages 8296--8306, 2023.

\bibitem[Xu et~al.(2024)Xu, Shi, Yifan, Chen, Yang, Peng, Shen, and Wetzstein]{xu2024grm}
Yinghao Xu, Zifan Shi, Wang Yifan, Hansheng Chen, Ceyuan Yang, Sida Peng, Yujun Shen, and Gordon Wetzstein.
\newblock Grm: Large gaussian reconstruction model for efficient 3d reconstruction and generation.
\newblock \emph{arXiv preprint arXiv:2403.14621}, 2024.

\bibitem[Yan et~al.(2024)Yan, Lin, Zhou, Wang, Sun, Zhan, Lang, Zhou, and Peng]{yan2024street}
Yunzhi Yan, Haotong Lin, Chenxu Zhou, Weijie Wang, Haiyang Sun, Kun Zhan, Xianpeng Lang, Xiaowei Zhou, and Sida Peng.
\newblock Street gaussians for modeling dynamic urban scenes.
\newblock \emph{arXiv preprint arXiv:2401.01339}, 2024.

\bibitem[Yang et~al.(2024)Yang, Kang, Huang, Zhao, Xu, Feng, and Zhao]{yang2024depth}
Lihe Yang, Bingyi Kang, Zilong Huang, Zhen Zhao, Xiaogang Xu, Jiashi Feng, and Hengshuang Zhao.
\newblock Depth anything v2.
\newblock \emph{arXiv preprint arXiv:2406.09414}, 2024.

\bibitem[Yu et~al.(2024{\natexlab{a}})Yu, Lu, Xu, Jiang, Xiangli, and Dai]{yu2024gsdf}
Mulin Yu, Tao Lu, Linning Xu, Lihan Jiang, Yuanbo Xiangli, and Bo Dai.
\newblock Gsdf: 3dgs meets sdf for improved rendering and reconstruction.
\newblock \emph{arXiv preprint arXiv:2403.16964}, 2024{\natexlab{a}}.

\bibitem[Yu et~al.(2024{\natexlab{b}})Yu, Sattler, and Geiger]{yu2024gaussian}
Zehao Yu, Torsten Sattler, and Andreas Geiger.
\newblock Gaussian opacity fields: Efficient and compact surface reconstruction in unbounded scenes.
\newblock \emph{arXiv preprint arXiv:2404.10772}, 2024{\natexlab{b}}.

\bibitem[Zhang et~al.(2025)Zhang, Bi, Tan, Xiangli, Zhao, Sunkavalli, and Xu]{zhang2025gs}
Kai Zhang, Sai Bi, Hao Tan, Yuanbo Xiangli, Nanxuan Zhao, Kalyan Sunkavalli, and Zexiang Xu.
\newblock Gs-lrm: Large reconstruction model for 3d gaussian splatting.
\newblock In \emph{European Conference on Computer Vision}, pages 1--19. Springer, 2025.

\bibitem[Zhang et~al.(2018)Zhang, Isola, Efros, Shechtman, and Wang]{zhang2018unreasonable}
Richard Zhang, Phillip Isola, Alexei~A Efros, Eli Shechtman, and Oliver Wang.
\newblock The unreasonable effectiveness of deep features as a perceptual metric.
\newblock In \emph{Proceedings of the IEEE conference on computer vision and pattern recognition}, pages 586--595, 2018.

\bibitem[Zhang et~al.(2024)Zhang, Ye, Chen, Chen, Zhang, Peng, Shi, and Zhao]{zhang2024drone}
Saining Zhang, Baijun Ye, Xiaoxue Chen, Yuantao Chen, Zongzheng Zhang, Cheng Peng, Yongliang Shi, and Hao Zhao.
\newblock Drone-assisted road gaussian splatting with cross-view uncertainty.
\newblock \emph{arXiv preprint arXiv:2408.15242}, 2024.

\bibitem[Zhao et~al.(2024)Zhao, Weng, Lu, Li, Li, Panda, and Xie]{zhao2024scaling}
Hexu Zhao, Haoyang Weng, Daohan Lu, Ang Li, Jinyang Li, Aurojit Panda, and Saining Xie.
\newblock On scaling up 3d gaussian splatting training.
\newblock \emph{arXiv preprint arXiv:2406.18533}, 2024.

\bibitem[Zhenxing and Xu(2022)]{zhenxing2022switch}
MI Zhenxing and Dan Xu.
\newblock Switch-nerf: Learning scene decomposition with mixture of experts for large-scale neural radiance fields.
\newblock In \emph{The Eleventh International Conference on Learning Representations}, 2022.

\bibitem[Zhou et~al.(2024)Zhou, Lin, Shan, Wang, Sun, and Yang]{zhou2024drivinggaussian}
Xiaoyu Zhou, Zhiwei Lin, Xiaojun Shan, Yongtao Wang, Deqing Sun, and Ming-Hsuan Yang.
\newblock Drivinggaussian: Composite gaussian splatting for surrounding dynamic autonomous driving scenes.
\newblock In \emph{Proceedings of the IEEE/CVF Conference on Computer Vision and Pattern Recognition}, pages 21634--21643, 2024.

\end{thebibliography}
}

\clearpage
\setcounter{page}{1}


\section{Data}
\label{supp:data}
\begin{figure}[htbp]
\centering
    \includegraphics[width=\linewidth]{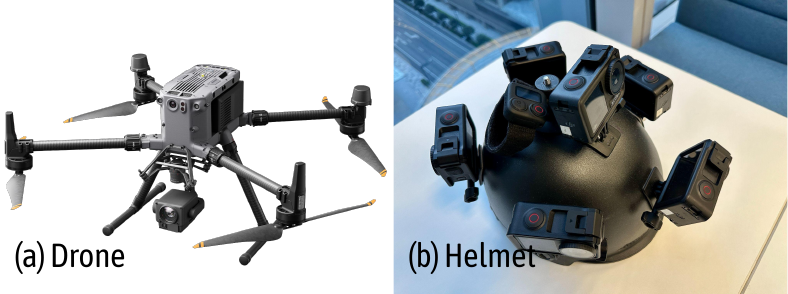}
    \vspace{-0.7cm}
    \caption{(a) M350RTK DJI drone for aerial images. (b) Helmet with six DJI Osmo Action4 cameras for street images}
    \vspace{-7pt}
    \label{fig:device}
\end{figure}

For real-world data, the aerial view images are captured by an M350RTK DJI drone equipped with five SHARE 304S cameras, as shown in Fig.~\ref{fig:device}(a). The resolution of these images is 9552 × 6368, and each camera has a sensor size of 36 mm.

Street view images are captured by a custom designed helmet equipped with six DJI Osmo Action4 cameras, following Hierarchical-3DGS~\cite{kerbl2024hierarchical}, as visualized in Fig.~\ref{fig:device}(b). The resolution of these images is 3840 × 2160. We use a DJI Osmo Action GPS Bluetooth remote to connect and operate all six cameras simultaneously. During the data collection process, we wear the helmet and walk to ensure image stability. The cameras are set to auto exposure, auto white balance, and timelapse mode with a 1-second interval. Each camera has a sensor size of 19.5 mm.

Following the setting of Gaussian Splatting~\cite{kerbl20233d}, we resize the the longest side original images to 1600 pixels. 

 For the synthetic data in our dataset,  we maintain the alignment of the cameras' roll, pitch, and yaw angles with those of the real-world scenes to ensure the uniformity of all data, as shown in Tab.~\ref{tab:rot_cams}.

\begin{table}[htbp]
    \centering
    \renewcommand{\arraystretch}{1.15}
    \setlength{\tabcolsep}{5pt}
    \begin{tabular}{c|ccc|ccc}
    \toprule
          & \multicolumn{3}{c|}{Aerial} & \multicolumn{3}{c}{Street} \\
        \multirow{-2}{*}{Rot}  & Roll & Pitch & Yaw  & Roll & Pitch & Yaw \\ \midrule
        1 & 0 & -45 & 0 & 0 & 0 & 0 \\
        2 & 0 & -45 & 90 & 0 & 25 & 0 \\
        3 & 0 & -45 & 180 & 0 & 0 & 75 \\
        4 & 0 & -45 & 270 & 0 & 0 & 145 \\
        5 & 0 & -90 & 0 & 0 & 0 & -145 \\
        6 &  &  &  & 0 & 0 & -75 \\
    \bottomrule
    \end{tabular}
    \caption{Camera rotation parameters in synthetic scenes.}
    \label{tab:rot_cams}
\end{table}

\section{More implementation}
\subsection{Global Appearance Embedding}
In large-scale scenes, the data is typically captured in different environments, leading to inconsistent exposures. Inspired by Octree-GS~\cite{ren2024octree} and Hierarchical-3DGS~\cite{kerbl2024hierarchical}, we employ classical generative Latent Optimization (GLO)~\cite{bojanowski2017optimizing} to optimize individual appearance embedding vectors for each training image. To ensure consistent appearance codes across different chunks, we initially train the Gaussian primitives without densification for a few iterations, as the appearance codes mainly capture global and low-frequency attributes of the scene.

\begin{table}[htbp]
\centering
\renewcommand{\arraystretch}{1.15}
\setlength{\tabcolsep}{2pt}
\resizebox{1\linewidth}{!}{
\begin{tabular}{l|ccc|ccc}
\toprule
\multicolumn{1}{c|}{\centering Scene} & \multicolumn{3}{c|}{Aerial} & \multicolumn{3}{c}{Street} \\  
\begin{tabular}{c|c} Method & Metrics \end{tabular}  & PSNR$\uparrow$ & SSIM$\uparrow$ & LPIPS$\downarrow$ & PSNR$\uparrow$ & SSIM$\uparrow$ & LPIPS$\downarrow$  \\
\midrule

Baseline~\cite{lu2023scaffold} & 20.18 & 0.539 & 0.549 & 21.22 & 0.626 & 0.394\\

Single Domain & \underline{22.42} & \underline{0.666} & \underline{0.402} & \underline{21.85} & \underline{0.653} & \underline{0.362}\\

Finetune  & 21.36 & 0.606 & 0.473 & 21.72 & 0.648 & 0.367\\

Ours  & \textbf{23.23} & \textbf{0.729} & \textbf{0.322} & \textbf{22.04} & \textbf{0.669} & \textbf{0.325} \\
\bottomrule
\end{tabular}}
\vspace{-6pt}
\caption{Quantitative comparison using naive finetuning solutions.}
\label{tab:naive_solution}
\vspace{-1em}
\end{table}

\begin{table*}[htbp]
\centering
\renewcommand{\arraystretch}{1.15}
\setlength{\tabcolsep}{1pt}
\resizebox{1\linewidth}{!}{
\begin{tabular}{l|ccc|ccc|ccc|ccc|ccc|ccc}
\toprule
 & \multicolumn{6}{c|}{City} & \multicolumn{6}{c|}{Colosseum} & \multicolumn{6}{c}{Elevenruin} \\ \cmidrule(l){2-19}
\multicolumn{1}{c|}{\multirow{-2}{*}{\centering Scene}} & \multicolumn{3}{c|}{Aerial} & \multicolumn{3}{c|}{Street} & \multicolumn{3}{c|}{Aerial} & \multicolumn{3}{c|}{Street} & \multicolumn{3}{c|}{Aerial} & \multicolumn{3}{c}{Street} \\  \midrule
\begin{tabular}{c|c} Method & Metrics \end{tabular}  & PSNR$\uparrow$ & SSIM$\uparrow$ & LPIPS$\downarrow$ & PSNR$\uparrow$ & SSIM$\uparrow$ & LPIPS$\downarrow$ & PSNR$\uparrow$ & SSIM$\uparrow$ & LPIPS$\downarrow$ & PSNR$\uparrow$ & SSIM$\uparrow$ & LPIPS$\downarrow$ & PSNR$\uparrow$ & SSIM$\uparrow$ & LPIPS$\downarrow$ & PSNR$\uparrow$ & SSIM$\uparrow$ & LPIPS$\downarrow$ \\
\midrule
2D-GS~\cite{huang20242d} & 25.27  & 0.739  & 0.391  & 21.75  & 0.708  & 0.439  & 22.50  & 0.752  & 0.382  & 25.76  & \cellcolor{tabthird}0.905  & 0.143  & 26.49  & 0.842  & 0.350  & 24.21  & 0.773  & 0.297  \\ 
Our-2D-GS & \cellcolor{tabsecond}32.21  & \cellcolor{tabsecond}0.931  & \cellcolor{tabsecond}0.113  & 23.94  & \cellcolor{tabthird}0.808  & \cellcolor{tabsecond}0.297  & \cellcolor{tabsecond}25.40  & \cellcolor{tabsecond}0.891  & \cellcolor{tabsecond}0.163  & \cellcolor{tabfirst}26.25  & 0.899  & 0.141  & \cellcolor{tabsecond}33.56  & \cellcolor{tabsecond}0.952  & \cellcolor{tabsecond}0.133  & \cellcolor{tabthird}26.12  & \cellcolor{tabthird}0.837  & \cellcolor{tabthird}0.211  \\ \midrule
3D-GS~\cite{kerbl20233d} & 26.79  & 0.784  & 0.351  & 21.79  & 0.723  & 0.422  & 22.25  & 0.754  & 0.380  & 25.30  & \cellcolor{tabsecond}0.910  & \cellcolor{tabsecond}0.132  & 27.49  & 0.857  & 0.333  & 24.87  & 0.795  & 0.276  \\ 
Scaffold-GS~\cite{lu2023scaffold} & \cellcolor{tabthird}30.03  & \cellcolor{tabthird}0.890  & \cellcolor{tabthird}0.187  & \cellcolor{tabthird}23.98  & 0.796  & 0.334  & \cellcolor{tabthird}25.14  & \cellcolor{tabthird}0.854  & \cellcolor{tabthird}0.226  & 25.33  & 0.867  & 0.187  & 31.21  & \cellcolor{tabthird}0.928  & \cellcolor{tabthird}0.175  & 26.10  & 0.835  & 0.219  \\ 
Hier-GS~\cite{kerbl2024hierarchical} & 29.15  & 0.871  & 0.206  & \cellcolor{tabfirst}24.51  & \cellcolor{tabsecond}0.810  & \cellcolor{tabthird}0.298  & 23.67  & 0.805  & 0.314  & \cellcolor{tabthird}25.74  & \cellcolor{tabfirst}0.915  & \cellcolor{tabfirst}0.129  & \cellcolor{tabthird}31.67  & 0.922  & 0.211  & \cellcolor{tabsecond}26.50  & \cellcolor{tabfirst}0.858  & \cellcolor{tabfirst}0.160  \\ 
Ours & \cellcolor{tabfirst}33.95  & \cellcolor{tabfirst}0.946  & \cellcolor{tabfirst}0.092  & \cellcolor{tabsecond}24.28  & \cellcolor{tabfirst}0.827  & \cellcolor{tabfirst}0.264  & \cellcolor{tabfirst}25.85  & \cellcolor{tabfirst}0.900  & \cellcolor{tabfirst}0.139  & \cellcolor{tabsecond}26.11  & 0.904  & \cellcolor{tabthird}0.133  & \cellcolor{tabfirst}34.99  & \cellcolor{tabfirst}0.967  & \cellcolor{tabfirst}0.071  & \cellcolor{tabfirst}26.67  & \cellcolor{tabsecond}0.855  & \cellcolor{tabsecond}0.173 \\ 
\bottomrule
\end{tabular}}
\vspace{-6pt}
\caption{Quantitative comparison on each synthetic scene of our proposed dataset.}
\label{tab:synthetic_result}
\vspace{-6pt}
\end{table*}

\subsection{Mesh Extraction}
For mesh extraction, we adopt the 2D-GS\cite{huang20242d} approach, rendering depth maps and fusing them into a TSDF volume, with the maximum depth range calculated based only on aerial views due to their wider coverage. The marching cube resolution is 1024.
 
\section{More Experiments}
\subsection{Naive Solutions}
\label{sec:naive_solution}

Based on the observations discussed in Section ~\ref{sec: challenges}, a naive solution is to merge the results from training on individual domains. To eliminate artifacts at the seams and maintain consistency in the feature space, we conduct an experiment where we concatenate the results from training on a single domain and fine-tuned the model for an additional 10k iterations on the Road and Park scenes. As shown in Tab.~\ref{tab:naive_solution}, this fine-tuning approach inefficient, time-consuming, and fails to address the core issue.

\subsection{More Quantitative Results}
We report quantitative results for each scene of our proposed dataset, as discussed in the main text: synthetic scenes (City, Colosseum, and Elevenruin) and real scenes (Road, Park). These results cover image quality metrics such as PSNR, SSIM~\cite{wang2004image}, and LPIPS~\cite{zhang2018unreasonable}, as shown in Tables~\ref{tab:synthetic_result},~\ref{tab:road},~\ref{tab:park}.

\begin{table}[htbp]
\centering
\renewcommand{\arraystretch}{1.15}
\setlength{\tabcolsep}{2pt}
\resizebox{1\linewidth}{!}{
\begin{tabular}{l|ccc|ccc}
\toprule
 & \multicolumn{6}{c}{Road} \\ \cmidrule(l){2-7}
\multicolumn{1}{c|}{\multirow{-2}{*}{\centering Scene}} & \multicolumn{3}{c|}{Aerial} & \multicolumn{3}{c}{Street} \\  \midrule
\begin{tabular}{c|c} Method & Metrics \end{tabular}  & PSNR$\uparrow$ & SSIM$\uparrow$ & LPIPS$\downarrow$ & PSNR$\uparrow$ & SSIM$\uparrow$ & LPIPS$\downarrow$  \\
\midrule
2D-GS~\cite{huang20242d} & 19.63  & 0.484  & 0.584  & 19.37  & 0.541  & 0.468  \\ 
Our-2D-GS & \underline{21.79}  & \underline{0.645}  & \underline{0.384}  & 20.57  & 0.628  & 0.349  \\ \midrule
3D-GS~\cite{kerbl20233d} & 19.95  & 0.509  & 0.562  & 20.17  & 0.573  & 0.435  \\ 
Scaffold-GS~\cite{lu2023scaffold} & 20.36  & 0.532  & 0.532  & 20.08  & 0.580  & 0.422  \\ 
Hier-GS~\cite{kerbl2024hierarchical} & 21.22  & 0.620  & 0.432  & \textbf{21.30}  & \textbf{0.651}  & \textbf{0.312}  \\ 
Ours & \textbf{22.60}  & \textbf{0.682}  & \textbf{0.356}  & \underline{20.94}  & \underline{0.637}  & \underline{0.341} \\ 

\bottomrule
\end{tabular}}
\vspace{-6pt}
\caption{Quantitative comparison on Road scene.}
\vspace{-1em}
\label{tab:road}
\end{table}

\begin{table}[htbp]
\centering
\renewcommand{\arraystretch}{1.15}
\setlength{\tabcolsep}{2pt}
\resizebox{1\linewidth}{!}{
\begin{tabular}{l|ccc|ccc}
\toprule
 & \multicolumn{6}{c}{Park} \\ \cmidrule(l){2-7}
\multicolumn{1}{c|}{\multirow{-2}{*}{\centering Scene}} & \multicolumn{3}{c|}{Aerial} & \multicolumn{3}{c}{Street} \\  \midrule
\begin{tabular}{c|c} Method & Metrics \end{tabular}  & PSNR$\uparrow$ & SSIM$\uparrow$ & LPIPS$\downarrow$ & PSNR$\uparrow$ & SSIM$\uparrow$ & LPIPS$\downarrow$  \\
\midrule
2D-GS~\cite{huang20242d} & 19.76  & 0.524  & 0.586  & 21.80  & 0.664  & 0.376  \\ 
Our-2D-GS & \underline{23.35}  & \underline{0.729}  & \underline{0.330}  & 22.46  & 0.681  & 0.339  \\ \midrule
3D-GS~\cite{kerbl20233d} & 20.23  & 0.545  & 0.565  & 22.64  & 0.681  & 0.361  \\ 
Scaffold-GS~\cite{lu2023scaffold} & 19.99  & 0.545  & 0.565  & 22.35  & 0.672  & 0.366  \\ 
Hier-GS~\cite{kerbl2024hierarchical} & 21.63  & 0.657  & 0.427  & \textbf{23.75}  & \textbf{0.720}  & \textbf{0.294}  \\ 
Ours & \textbf{23.85}  & \textbf{0.776}  & \textbf{0.287}  & \underline{23.14}  & \underline{0.701}  & \underline{0.308} \\ 

\bottomrule
\end{tabular}}
\vspace{-6pt}
\caption{Quantitative comparison on Park scene.}
\vspace{-1em}
\label{tab:park}
\end{table}

\subsection{Ablation}
\label{supp:ablation}
We select Scaffold-GS~\cite{lu2023scaffold} as our baseline and perform two additional ablation studies focusing on the fine stage and global appearance embedding, respectively. For quantitative results, we use the Road and Park scenes, while Block\_A is used for qualitative analysis.

\begin{table}[htbp]
\centering
\renewcommand{\arraystretch}{1.15}
\setlength{\tabcolsep}{3pt}
\resizebox{1\linewidth}{!}{
\begin{tabular}{l|ccc|ccc}
\toprule
\multicolumn{1}{c|}{\centering Scene} & \multicolumn{3}{c|}{Aerial} & \multicolumn{3}{c}{Street} \\  
\begin{tabular}{c|c} Method & Metrics \end{tabular}  & PSNR$\uparrow$ & SSIM$\uparrow$ & LPIPS$\downarrow$ & PSNR$\uparrow$ & SSIM$\uparrow$ & LPIPS$\downarrow$  \\
\midrule

Baseline~\cite{lu2023scaffold} & 20.18 & 0.539 & 0.549 & 21.22 & 0.626 & 0.394 \\
Ours w/o fine stage & \textbf{23.32} & \underline{0.725} & \underline{0.326} & 21.69 & 0.658 & 0.338 \\
Ours  & \underline{23.23} & \textbf{0.729} & \textbf{0.321} & \textbf{22.04} & 0.669 & 0.324 \\

\bottomrule
\end{tabular}}
\vspace{-6pt}
\caption{Ablations on our proposed real-world scenes.}
\label{tab:ablate_supp}
\vspace{-1em}
\end{table}

\paragraph{Fine Stage.} The second stage is used for complementing the details of aerial views. The rendering quality will decrease hugely if discarding it, as shown in Tab.~\ref{tab:ablate_supp}.


\section{Limitation and More Discussion}
While our method proves effective in reconstructing large-scale aerial-to-ground scenes and producing high-quality results, it also has certain limitations. First, similar to most Gaussian-based methods, \modelname may reach suboptimal solutions when there is insufficient input information. In future work, we plan to leverage advanced foundation models to guide the optimization process more effectively.
Additionally, the divide-and-conquer approach inevitably introduces redundancy due to the required overlaps for seamless merging between chunks. Implementing more systematic approaches, such as Grendel-GS~\cite{zhao2024scaling} or RetinaGS~\cite{li2024retinags}, also presents a promising solution.


\end{document}